\documentclass[a4paper,fleqn]{cas-dc}

\usepackage[authoryear]{natbib}
\usepackage[english]{babel}
\usepackage{amsthm}
\usepackage{amsmath,amsfonts,bm}
\usepackage{optidef}
\usepackage{leftindex}
\usepackage{commath}
\usepackage{cancel}
\usepackage{siunitx}
\usepackage{microtype}
\usepackage[ruled,vlined]{algorithm2e}
\usepackage[normalem]{ulem}
\usepackage[export]{adjustbox}
\usepackage{stfloats}
\usepackage{subfig}
\usepackage{orcidlink}
\usepackage{hyperref}

\newtheorem{thm}{Theorem}

\newtheorem{proposition}{Proposition}
\newtheorem{remark}{Remark}

\def\wrt{w.r.t.\@\xspace}
\def\ie{i.e.\@\xspace}
\def\eg{e.g.\@\xspace}

\def\tsc#1{\csdef{#1}{\textsc{\lowercase{#1}}\xspace}}
\tsc{WGM}
\tsc{QE}

\ExplSyntaxOn
\bool_gset_true:N \g_stm_nologo_bool   
\ExplSyntaxOff
\RenewDocumentCommand \printorcid { }
  {
  }
\ExplSyntaxOn
\cs_new_protected:Npn \__arxiv_strip_orcid:n #1
  {
    \tl_set:Nn \l_tmpa_tl { #1 }
    \regex_replace_all:nnN { \c{orcidlink}\{.*?\} } { } \l_tmpa_tl
    \regex_replace_all:nnN { \c{textsuperscript}\{\} } { } \l_tmpa_tl
    \tl_use:N \l_tmpa_tl
  }
\RenewDocumentCommand \creditauthor { m m }
  {
    \seq_gput_right:Nn \g_stm_credits_seq
      { { \exp_not:N\bfseries \__arxiv_strip_orcid:n { #2 } : }\space #1 }
  }
\RenewDocumentCommand \emailauthor { m m }
  {
    \int_gincr:N \g_ead_int
    \tl_set:Nn \l_tmpa_tl { #2 }
    \regex_replace_all:nnN { \c{orcidlink}\{.*?\} } { } \l_tmpa_tl
    \regex_replace_all:nnN { \c{textsuperscript}\{\} } { } \l_tmpa_tl
    \seq_gput_right:Nx \g_stm_ead_seq
      {
        { \exp_not:n { \ttfamily \tl_to_str:n { #1 } } }
        \exp_not:n { \parsename } { \exp_not:V \l_tmpa_tl }
        \exp_not:n { \space(\eadauthor) }
      }
  }
\ExplSyntaxOff

\ExplSyntaxOn
\cs_set:Npn \__first_footerline: { \hbox{} }
\ExplSyntaxOff

\usepackage{etoolbox,xpatch}
\newcommand{\arxivnotice}{%
  \par\small\centering
  This is the accepted manuscript of an article published in
  \textit{Control Engineering Practice}, 176 (2026), 107166.
  The version of record is available at
  \url{https://doi.org/10.1016/j.conengprac.2026.107166}.
  \copyright{} 2026.
  This manuscript version is made available under the
  CC~BY-NC-ND~4.0 license:
  \url{https://creativecommons.org/licenses/by-nc-nd/4.0/}.
  \dashrule{0pt}{3pt}}
\xpretocmd{\Abstract}{\arxivnotice}{}{}

\begin{document}
    \let\WriteBookmarks\relax
    \def\floatpagepagefraction{1}
    \def\textpagefraction{.001}
    \shorttitle{Reinforced Iterative Learning Control for elastic joint arms}
    \shortauthors{Y. De~Santis et~al.}

    \title [mode = title]{Reinforced Iterative Learning Control for elastic joint arms:\\ Guaranteed cross-trajectory generalization under model uncertainties}
    \tnotemark[1]

    \tnotetext[1]{This document is the result of the research
    project funded by the Italian Ministry of University and Research (MUR) - Fondo Italiano per la Scienza Applicata (FISA), through the OCCAM project under Grant FISA 2023-00324 CUP I53C25000700001 and the ``FoReLab'' (Future-oriented Research Lab) Project (Departments of Excellence), and partially by the European Union HORIZON-MSCA-2023-SE-01-01 - MSCA Staff Exchanges 2023 Program under the Grant Agreement No.~101182891 (NEUTRAWEED).}

    \author[1,2]{Y. De~Santis\textsuperscript{\orcidlink{0009-0007-6161-3141}}}[%
        type=editor,
        auid=000,
        bioid=1,
        ]
    \cormark[1]
    \fnmark[1]
    \ead{yuri.desantis@phd.unipi.it}
    \credit{Writing - review \& editing, Writing - original draft, Conceptualization, Software, Methodology, Validation}
    \author[1]{M. Pierallini\textsuperscript{\orcidlink{0000-0003-0547-2747}}}[%
        type=editor,
        auid=000,
        bioid=2,
        ]
    \credit{Writing - review \& editing, Conceptualization, Methodology}
    \author[1,2]{F. Angelini\textsuperscript{\orcidlink{0000-0003-2559-9569}}}[%
        type=editor,
        auid=000,
        bioid=3,
        ]
    \credit{Writing - review \& editing, Conceptualization, Supervision}
    \author[1,2]{M. Garabini\textsuperscript{\orcidlink{0000-0002-5873-3173}}}[%
        type=editor,
        auid=000,
        bioid=4,
        ]
    \credit{Writing - review \& editing, Supervision, Project administration, Funding acquisition}
    \affiliation[1]{organization={Centro di Ricerca ``En\-ri\-co Pi\-ag\-gio'', U\-ni\-ver\-si\-t\`{a} di Pisa},
                    addressline={Largo Lucio Lazzarino 1},
                    city={Pisa},
                    postcode={56122},
                    country={Italy}}

    \affiliation[2]{organization={Dipartimento di Ingegneria dell'Informazione, U\-ni\-ver\-si\-t\`{a} di Pisa},
                    addressline={Via G. Caruso 16},
                    city={Pisa},
                    postcode={56122},
                    country={Italy}}

    \cortext[cor1]{Corresponding author}

    \begin{abstract}
Trajectory tracking for compliant robots is challenging due to nonlinear dynamics and modeling uncertainties, often requiring learning-based methods to avoid explicit parameter identification. 
Iterative Learning Control (ILC) ensures theoretically high precision and preserves the passive compliance of the robot. However, traditional ILC requires a separate learning phase for each new trajectory.
Conversely, Reinforcement Learning (RL) generalizes to new trajectories, but practically suffers from high sample complexity and limited tracking precision.
This work proposes Reinforced Iterative Learning Control (RILC) that combines the strengths of both approaches to achieve fast adaptation and guaranteed tracking convergence across diverse trajectories. By establishing a concurrent coupling between the two controllers, RILC overcomes the high initial errors and slow adaptation of traditional ILC, enabling high-performance tracking from the very first trials. In this architecture, 
ILC provides high-quality samples, while RL contributes as a generalizing best-effort policy across iterations. Additionally, a theoretical proof is presented to show that the RL component preserves ILC stability and, under specific conditions, can also reduce the error bound asymptotically. 
The efficacy of this framework is demonstrated through simulation and experiments on an underactuated 4-degree-of-freedom serial elastic robot using different trajectories and modeling mismatches.
\end{abstract}


    \maketitle

    \section{Introduction}
The recent successes of robots deployed in unstructured environments, such as environmental monitoring 
\cite{pierallini2025fishing}, and robot-human interactions \cite{li2020skill} emphasize the scientific community's interest in robots that operate with limited information. Furthermore, enhancing performance and safety involves compliant hardware elements to guarantee physical safety \cite{angelini2018decentralized} and learning-based controllers~\cite{peters2016robot}.
However, the intrinsic flexibility of these structures introduces complex nonlinear dynamics that make trajectory tracking control challenging \cite{de2016robots, shao2023model}.

Model-based control ensures performance but requires precise models, which are difficult to obtain for compliant systems \cite{de2016robots,george2018control}. To handle modeling errors, high feedback gains are often used, effectively adding a virtual stiffness in parallel with the physical transmission and yielding an increased closed-loop stiffness that negates the safety advantages of the mechanical adaptability \cite{angelini2018decentralized}. This is particularly critical for collaborative robots with elastic elements and industrial arms with parasitic flexibility such as harmonic drives or belt transmissions \cite{de2016robots}. While data-driven identification can address these uncertainties, it remains costly and time-consuming \cite{hollerbach2016model}.

Learning techniques obtain reliable models or data across diverse dynamic scenarios, see \cite{sutton2018reinforcement, pierallini2023iterative}. This approach can update and reshape the control directly from raw real-time data in a model-free and data-driven fashion. 
However, data-driven approaches are time-consuming and need a large amount of data.
The combination of MB and data-driven control is a promising choice for obtaining good results with guarantees of MB while adding an auxiliary part with learning.

One example of learning controllers is Iterative Learning Control (ILC) \cite{ahn1993iterative, bristow2006survey, chen2020generalized, pierallini2021robust, aarnoudse2024design}, which guarantees theoretically good performance using mostly feedforward actions while compensating for the reality gap by repeating the desired task, iteratively. Recent research has investigated multi-reference tracking through various generalization strategies, including parametric basis functions \cite{blanken2016batch}, combined trial-specific adaptations \cite{tsurumoto2023task, tsurumoto2024combined}, and non-parametric Gaussian Process methods \cite{meindl2024ai, sun2024parameterised}. However, achieving fast adaptation across diverse trajectories remains a challenge, because these methods often restrict target trajectories to a narrow neighborhood of the nominal path to preserve stability. 
Conversely, Reinforcement Learning (RL) generalizes across trajectories, enabling zero-shot transfer to new references \cite{kaelbling1996reinforcement, schulman2017proximal}, but suffers from high sample complexity and policy chattering that limits tracking precision \cite{kober2013reinforcement, pavlichenko2022real}.
While the similarities and differences in ILC and RL learning processes have generated interest \cite{zhang2019preliminary}, existing combinations treat the two controllers as functionally decoupled. Some employ additive blending with fixed weights \cite{xu2020iterative, he2025iterative} or RL-based parameter tuning \cite{soleimani2025data}, while others adopt sequential pipelines where ILC generates offline datasets for RL training \cite{ma2023reinforcement, yang2026reinforcement}. None of these approaches provide formal convergence guarantees for the combined system, and the RL component remains blind to the trial-to-trial evolution of the ILC feedforward signal. Other works modify the RL algorithm itself by drawing inspiration from ILC, \eg, \cite{gurumurthy2023deep}. Our objective, however, is not to improve the RL algorithm internally, but rather to operate directly at the control level.

\begin{figure}
    \centering
    \includegraphics[width=0.9\linewidth]{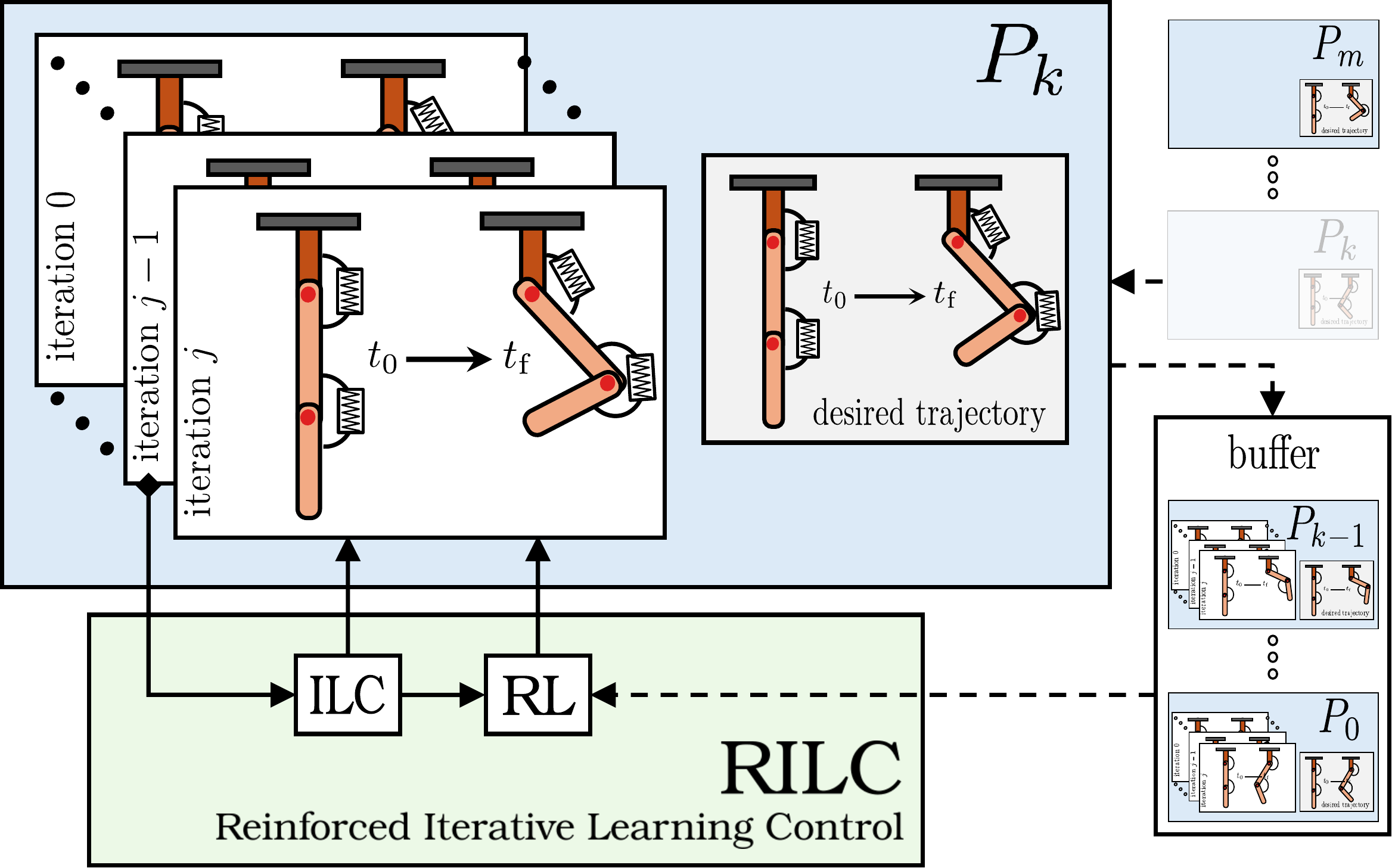}
    \caption{Reinforced Iterative Learning Control (RILC) scheme. The RILC combines Reinforcement Learning (RL) and Iterative Learning Control (ILC) terms (green block). 
    Each process (blue blocks) is populated by the iterations required for convergence to a predefined desired trajectory (grey block).
    The ILC is updated using past iterations $0, \cdots, j-1$, while the RL control is determined by the current ILC action and information gathered from the current process $P_{k}$. A buffer stores previously executed processes $0, \cdots, P_{k-1}$, and it is used to update the RL policy.
    }
    \label{fig:general_ctrl_scheme}
\end{figure}

This paper proposes a new control framework, Reinforced Iterative Learning Control (RILC), which combines the ILC and RL contributions to achieve precise multi-reference tracking. RILC establishes a concurrent coupling where the RL policy observes the ILC historical states and actively shapes the feedforward signal in real-time, while the ILC loop provides a stable envelope that constrains RL exploration. This synergy overcomes the slow trial-by-trial adaptation of traditional ILC, enabling high-performance execution from the very first trials. As a practical benefit, accurate feedforward compensation reduces the reliance on aggressive active feedback, which is essential in collaborative robotics and contact-rich industrial tasks to maintain the hardware's safety and mechanical adaptability.
We derive theoretical convergence guarantees and conditions that ensure tighter error bounds than baseline ILC. 

We validate the effectiveness of the algorithm both in numerical simulations, to test robustness against strong modeling mismatches, and in real-world experiments using a 4-degree-of-freedom (DoFs) articulated soft robot \cite{pratt1995series}. Results confirm the robustness of RILC on complex physical dynamics, outperforming baseline ILC and RL controllers.

This work is organized as follows. In Section \ref{sec:state}, we introduce the state-of-the-art learning frameworks used; in Section \ref{sec:p_def}, we describe the control problem in a model-based and model-free paradigm; in Section \ref{sec:p_sol} we derive the control architecture, and show how all the introduced issues can be addressed. In Section \ref{sec:val}-\ref{sec:disc}, the controller effectiveness and robustness are tested and discussed. Finally, in Section \ref{sec:concl}, we conclude.

    \section{State-of-the-art}
\label{sec:state} This section reports the state-of-the-art of ILC, RL, and
their combined approach.

\subsection{Iterative Learning Control (ILC)} \label{sec:ilc} 
ILC has been successfully applied to tracking tasks across diverse domains, from chemical processes, \cite{mezghani2002application}, and drones, \cite{nguyen2021iterative}, to various industrial platforms, \cite{axehill2014estimation, chen2020generalized} and soft robotics, \cite{angelini2018decentralized}. Traditional convergence and stability guarantees in ILC are obtained via model-based paradigms, \cite{wei2016output, shen2020zero, pierallini2021robust, aarnoudse2024design}.

For elastic joint systems, the literature is broadly divided into non-collocated and collocated configurations, see \cite{de2016robots}. When link-side variables are unmeasured, non-collocated approaches must integrate state and disturbance observers \cite{talole2009extended, rsetam2019cascaded, yan2023time}, or sensor fusion techniques \cite{axelsson2012bayesian, axehill2014estimation}, to estimate the link states for the control loop. However, this introduces sensitivity to estimation errors, which can degrade tracking performance. On the other hand, collocated configurations avoid estimation-based control, but must still compensate for the joint deflection reflecting across the transmission. For these setups, ILC is particularly suitable because it operates mainly in a feedforward fashion, often combined with low-gain feedback loops to preserve the robot's physical compliance \cite{angelini2018decentralized, pierallini2023iterative}. To systematically optimize the feedforward action while handling system constraints, the standard formulation \cite{ahn1993iterative} is extended to Norm-Optimal ILC (NOILC) \cite{amann1996iterative}, which guarantees monotonic convergence by exploiting model information. However, its effectiveness is strictly limited by the accuracy of the model and computational complexity, \cite{jiang2022norm}.

To extend traditional ILC beyond single-reference tracking, the focus has shifted toward multi-reference learning frameworks. On one hand, parametric methods employ basis functions \cite{blanken2016batch} to project updates onto trajectory primitives for linear systems. These have evolved into combined designs \cite{tsurumoto2023task, tsurumoto2024combined} that merge shared primitives with trial-specific adaptations considering their mutual interaction. However, when operating sufficiently far from the training trajectories, their tracking performance degrades, reverting to standard single-reference ILC behavior. On the other hand, non-parametric methods such as Kernel-based ILC and Gaussian Process Regression have been explored \cite{meindl2024ai, sun2024parameterised}. 
While these approaches achieve precise trajectory-tracking for points sufficiently close to the training datasets, even in the presence of nonlinearities, the former risks overfitting sparse data, whereas the cubic computational scaling of the latter severely restricts real-time applicability.

Despite these advancements, achieving fast adaptation across diverse trajectories for elastic joint systems remains an open challenge. Multi-reference ILC requires frameworks capable of scaling to large workspaces in the presence of strong nonlinearities while preserving both fast initial adaptation and asymptotic convergence.

\subsection{Reinforcement Learning (RL)}
\label{sec:rl} 

RL has achieved great success in solving robot control problems without requiring an explicit analytical model \cite{kober2013reinforcement, rudin2022learning, esser2022guided}.
For trajectory tracking, online model-free algorithms, such as Proximal Policy Optimization (PPO) \cite{schulman2017proximal}, parameterize the control policy to map system observations to joint torques or references. The strength of this formulation is that the learned policy can be deployed on new trajectories within the training distribution without re-tuning, demonstrating effective zero-shot transfer capability.

For systems with elastic elements, RL has been explored in contexts such as trajectory tracking on flexible-joint manipulators \cite{pavlichenko2022real}, force control on series elastic actuators \cite{sambhus2023real}, and manipulation with soft robotic arms \cite{alessi2024pushing}. These works confirm that RL can learn state-aware strategies on compliant hardware without explicit model identification.
Moreover, they show that the policy alone is insufficient for high-precision tracking, requiring external correction layers \cite{pavlichenko2022real}, filtering structures \cite{sambhus2023real}, or supplementary control layers \cite{alessi2024pushing} to achieve acceptable performance.
The inherent nature of deep neural network (DNN) function approximation leaves a persistent tracking error, motivating hybrid architectures where a nominal controller handles base dynamics while RL compensates for the residual dynamics \cite{johannink2019residual}.
High tracking performance therefore relies on collecting informative samples, often requiring strategies like curriculum learning \cite{narvekar2020curriculum} to cover the relevant regions of the state space. However, on elastic joint systems the link-motor coupling increases the complexity of the learning task, and the sample complexity grows accordingly \cite{kober2013reinforcement}. This curse of dimensionality makes data collection on physical compliant hardware both time-consuming and mechanically demanding.

Practical deployment faces the sim-to-real gap \cite{mehta2020active}, arising from system mismatches. Standard RL policies use frozen weights and cannot correct residual errors on unseen states. Online adaptation is typically constrained by sample inefficiency and the risk of catastrophic forgetting, which can compromise the policy's generalization capabilities \cite{kober2013reinforcement}.
These aspects reveal an inherent gap where RL, despite providing a generalizing policy, typically offers only probabilistic convergence and requires informative samples to reduce residual errors. This motivates combining RL with a complementary framework that generates high-quality data and structures observations to keep the policy within its training distribution, ensuring closed-loop stability.

\subsection{Combined approaches of ILC and RL}
\label{sec:rilc} 
The similarities and differences between the learning mechanisms of ILC and RL, particularly how they tackle trajectory tracking and use past samples to optimize control, have generated interest, \cite{zhang2019preliminary}.

To enhance the tracking performance of ILC and extend its applicability beyond strictly repetitive tasks, several works combine ILC with RL.
One approach uses RL to enhance the capability of ILC to handle non-repetitive disturbances and sudden trajectory variations. While standard ILC relies on strict repeatability, these methods employ RL to compensate when this assumption is violated. For instance, in \cite{soleimani2025data}, a policy selects a forgetting factor from a discrete set to weight past control inputs, allowing ILC to maintain its classical structure while handling iteration-varying uncertainties. The method provides theoretical convergence guarantees, yet validation is restricted to simulation on linear systems. 
Other approaches integrate the RL action directly into the control loop alongside ILC. In \cite{xu2020iterative}, a fixed blending factor combines a P-type ILC output with an RL policy action to handle dramatic trajectory changes in batch processes. However, the role and evolution of this blending factor are not discussed, which is a critical aspect for formal theoretical analysis that is missing. In \cite{he2025iterative}, RL is integrated within a model-based ILC framework. Unlike simple blending, this approach uses ILC as a stable baseline, and the RL agent acts as an adaptive compensator to handle non-repetitive disturbances and model mismatches. While these works enhance robustness for a single task, they are often validated only in simulation and do not address the core problem of generalization to new reference trajectories. Since the RL component operates independently from the ILC evolution across iterations, it cannot anticipate or compensate for the trial-to-trial changes in the feedforward signal.
To achieve generalization across diverse trajectories, other methods decouple the learning phase from execution. In \cite{ma2023reinforcement}, the authors reformulates the RL problem as a supervised learning task, using ILC offline as a data generator to train a neural network that maps desired trajectories to optimal feedforward inputs. Experiments on a robotic system with elastic parts demonstrates transfer to new target locations, but the learning phase remains entirely offline, preventing real-time adaptation. In \cite{yang2026reinforcement}, an RL agent is pre-trained to imitate ILC to reduce exploration costs, followed by online fine-tuning on a nonlinear system. However, the pre-training relies on a heuristic switching logic between controllers, which limits the RL policy from developing an independent optimal strategy. These methods assume that offline data given to RL policy covers the relevant state space, which prevents the ability to compensate for the reality gap. Furthermore, neither work provides formal convergence guarantees for the combined framework.

Existing methods isolate one controller from the other through parameter adjustment, additive superposition, or temporal decoupling. Consequently, these approaches often rely on the assumption that the RL agent does not destabilize the base loop, lacking formal proofs of combined convergence. Therefore, developing a concurrent RL-ILC framework that ensures joint stability, enables fast adaptation across iterations, and achieves generalization on compliant systems remains an open challenge.
%
%

    \section{Preliminaries and problem statement}
\label{sec:p_def}

In this section, we first present the notation used, the representation of robotic systems, and the optimal control formulation in a model-free approach to tackle the trajectory tracking problem. Finally, we report the goals of this work.

\subsection{Notation}
This subsection aims to align the notation and terminology used in this paper.
Although the use of RL framework, we prefer to use optimal control theory terminology where it is possible. For this reason, we approach the problem as a minimization of the functional cost instead of a maximization of the cumulative reward. Bearing in mind that, we consider $c(\cdot) = - r(\cdot)$, where $c(\cdot)$ and $r(\cdot)$ are the running cost and the immediate reward, respectively.
We call the training phase the updating phase of RL parameters during simulation experiments.

Let $z(\cdot) : [0, t_\textnormal{f}]\to \mathbb{R}^{n}$ be a vector function, we denote with $z^{(i)}(t)$ its $i$-th time derivative, with $\Vert{z}\Vert_\lambda$, the $\lambda$-norm of a vector, \ie $\Vert{z}\Vert_\lambda \leq \sup_t\{\Vert{z}\Vert e^{-\lambda t}\}$ for $\lambda > 0$, with $\check{z}(\cdot)$ a sample of $z(\cdot)$, finally, with $\hat{z}(\cdot)$ an approximation of $z(\cdot)$.

Let $z(\cdot): [0, t_\textnormal{f}]\to \mathbb{R}^{n}$ be a discrete time vector function, $[\cdot]_+$ is the discrete-time shift operator, \ie, $z(k)_+=z(k+1)$.

Finally, for the sake of space, we will omit the explicit time dependence where it is trivial. Refer to Table~\ref{tab:notation} for a summary of the main notation.

\begin{table}[ht]
\centering
\caption{Principal notation used throughout the paper.}
\label{tab:notation}
\small
\begin{tabular}{ll}
\hline
\textbf{Symbol} & \textbf{Description} \\
\hline
$j$ & Iteration index \\
$x_j$ & State vector at iteration $j$ \\
$y_j$ & Output \\
$e_j$ & Tracking error $y_d - y_j$ \\
$\bar{u}_j$ & ILC feedforward control input \\
$\tilde{u}_\nu(\xi_j)$ & RL control contribution, parametrized by $\nu$ \\
$u_{\textnormal{mb}}$ & Model-based feedforward control \\
$u_{\textnormal{fb}}$ & Low-gain feedback control \\
$\delta\bar{u}_j$ & ILC input error $\bar{u}_d - \bar{u}_j$ \\
$\delta\tilde{u}_j$ & RL input error $\tilde{u}_\nu(\xi_d) - \tilde{u}_\nu(\xi_j)$ \\
$\xi_j$ & Measurable process variable (observation) \\
$\chi_i$ & Gains for the error derivatives $e_j^{(i)}$ \\
$\mathfrak{L}_j$ & Iteration-dependent ILC learning operator \\
$\Omega$ & RILC contraction rate \\
\hline
\end{tabular}
\end{table}

\subsection{Model formulation}
Let us consider the square system of a $2n$-DoFs under-actuated by Series Elastic Actuator (SEA) compliant systems, as derived in \cite{albu2016actuators}, \ie,
\begin{equation}
\label{eq:sys_lagrange}
    \left\{\begin{array}{l}
    M (q)\ddot{q} + C(q, \dot{q}) \dot{q} + G(q) + K(q-\theta) = 0 \\
    B\ddot\theta + D\dot\theta - K(q-\theta) = u \,,
    \end{array}\right.
\end{equation}
where $q, \dot{q}, \ddot{q} \in \mathbb{R}^{n}$ and $\theta, \dot{\theta}, \ddot{\theta} \in \mathbb{R}^{n}$ are the position, velocity, and acceleration vectors on the link-side and motor-side, respectively. $M(q) \in \mathbb{R}^{n \times n}$ is the robot inertia matrix, $C(q, \dot{q}) \in \mathbb{R}^{n \times n}$ is the robot Coriolis matrix, $G(q) \in \mathbb{R}^{n}$ is the gravity vector, $K \in \mathbb{R}^{n \times n}$ is the stiffness matrix associated to SEA. $B,D \in \mathbb{R}^{n \times n}$ are the motor inertia matrix and damping matrix $D \succ 0$, respectively. $u \in \mathbb{R}^{n}$ is the torque control input. 
Let us consider a set $\mathcal{P}=\{P_1, \ldots, P_k, \ldots P_m\}$ where $P_k$ represents the $k$-th iterative process, for each iterative process $P_k$ consider the iterative process with index $j \in \mathbb{N}_0$.
We follow the classic affine state-space form defining the state $x_j\doteq [q_j^\top, \theta_j^\top,  \dot{q}_j^\top, \dot{\theta}_j^\top]^\top \in \mathbb{R}^{4n}$ for continuous-time systems with $t \in\left[0, t_{\textnormal{f}}\right]$ is the time variable ($t_{\textnormal{f}}$ terminal time):
\begin{equation}
\label{eq:sys_eq}
    \left\{\begin{array}{l}
    \dot{x}_{j}=f(x_{j})+g (x_{j}) \left[\bar{u}_{j} + \tilde{u}_\nu(\xi_j)\right]  \\
    y_{j}=h (x_{j}) \\
    \xi_j =\phi(y_\textnormal{d}, y_{j-I}, \bar{u}_{j-I}) \,,
    \end{array}\right. 
\end{equation}
with $x_{j}(t=0) \in \mathbb{R}^{4 n}$ as initial condition:
$\bar{u}_{j}$, $\tilde{u}_\nu(\xi_j) \in \mathbb{R}^{n}$ are the control inputs, i.e., $u_j=\bar{u}_{j} + \tilde{u}_\nu(\xi_j) $, where $\tilde{u}(\cdot)$ is parametrized by $\nu$, $\xi_j(\cdot)$ is defined by $\phi(\cdot)$ that is a function of measurable process variables, $I\subseteq\{0,\ldots,j\}$ and we denote $z_{j-I}=\{z_{j-i}\}, \forall i\in I$. 
$y_{j} \in \mathbb{R}^{n}$ is the output, $h(\cdot): \mathbb{R}^{4 n} \times[0, t_{\textnormal{f}}] \rightarrow \mathbb{R}^{n}$ is the output function defined as $h(x_{j}) = S_\theta x_j$, where $S_\theta = [0_{n \times n}\,, \,I_{n}\,, \,0_{n \times 2n}] \in \mathbb{R}^{n \times 4n} $ is the selection matrix for $\theta$ in $x$.
$f(\cdot): \mathbb{R}^{4 n} \times[0, t_{\textnormal{f}}] \rightarrow \mathbb{R}^{4n}, g(\cdot): \mathbb{R}^{4 n} \times$ $[0, t_{\textnormal{f}}] \rightarrow \mathbb{R}^{4 n \times n}$ are the drift and control vector fields
\begin{align}
    f (x_{j})&=
    \left[\begin{array}{c}
        \dot{q}_{j} \\
        \dot{\theta}_{j} \\
        - N_1(q_{j}, \theta_{j}, \dot{q}_{j})\\
        - N_2(q_{j}, \theta_{j}, \dot{\theta}_{j})
    \end{array}\right] , \;
    g (x_{j}) =
    \left[\begin{array}{c}
    0_{3n \times n}\\ 
    B^{-1}
    \end{array}\right] \,,
\end{align}
with $N_1(q_{j}, \theta_{j}, \dot{q}_{j}) \doteq M^{-1}(q_{j}) [C(q_{j}, \dot{q}_{j}) \dot{q}_{j} + G(q_{j}) + K(q_{j}-\theta_{j})]$ and $N_2(q_{j}, \theta_{j}, \dot{\theta}_{j}) \doteq B^{-1}[D\dot{\theta}_{j}-K(q_{j}-\theta_{j})]$.

Note that for each $P_k$ is associate to a particular desired trajectory to follow $y_{\textnormal{d}}(t):\left[0, t_{\textnormal{f}}\right] \rightarrow \mathbb{R}^{n}$. We denote $y_{\textnormal{d}}(t)$ of the process $P_k$ as $\leftindex^k y_{\textnormal{d}}(t)$. We omit the left index $k$ for simplicity when its dependence is trivial.
We assume that:
\begin{itemize}
\label{item:assump}
    \item[A1)] The system \eqref{eq:sys_eq} has a fixed vector relative degree $r_{\textnormal{v}}$, $\forall P_k \in \mathcal{P}$, $\forall x \in \mathbb{R}^{4 n}$  $r_{\textnormal{v}}=[r_{1}, \ldots, r_{n}]$  $\forall t \in [0, t_\textnormal{f} ]$, see, e.g., \cite{isidori1995elementary}. Furthermore, we assume that $r=r_{1}=\cdots=r_{n}> 0$.  The matrix $E(x) \in \mathbb{R}^{n \times n}$ is computed such as $E_{i, s}(x) \doteq$ $L_{g_{i}} L^{r_{s-1}} h_{s}(x), \forall i, s=1, \ldots, n$ and is commonly called the decoupling matrix of the system \eqref{eq:sys_eq}. \label{a1}
    
    \item[A2)] The vector fields $f(\cdot), L_{f}^{i} h(\cdot), i=0, \ldots, r$, $h(\cdot), g(\cdot)$, and $E(\cdot)$ are globally Lipschitz with unknown constants $f_{0}, \psi_{0}^i,$ for $i=0, \ldots, r, h_{0}, g_{0}$, and $E_{0}$ respectively, i.e., $\Vert f(\bar{x}_{1})-f(\bar{x}_{2})\Vert \leq f_{0}\Vert\bar{x}_{1}-\bar{x}_{2}\Vert $, $\forall \bar{x}_{1}, \bar{x}_{2} \in \mathbb{R}^{4 n}, $ and $f_{0} \in \mathbb{R}\backslash\{+\infty\}, \forall P_k \in \mathcal{P}$.\label{a2}
    
    \item[A3)] The desired output trajectory $y_{\textnormal{d}}(t):[0, t_{\textnormal{f}}] \rightarrow$ $\mathbb{R}^{n}$ is feasible, continuous, bounded, and differentiable for, at least, $r$ times, $\forall t \in[0, t_{\textnormal{f}}], \forall P_k \in \mathcal{P}$.\label{a3}

    \item[A4)] The initial condition $x_j(0) \in \mathbb{R}^{4n}$ presents a bounded shift from the desired one, $\forall P_k \in \mathcal{P}$, i.e., $x_j (0) = x_{\textnormal{d}}(0) + l_j$, $l_j \in \mathbb{R}^{4n}$, $\forall j \in \mathbb{N}_0$, i.e., $\sup_j \{\Vert{l_j}\Vert \} \leq b_\textnormal{l}$, $b_\textnormal{l} \in \mathbb{R} \backslash\{+\infty\}$.\label{a4}
    
    \item[A5)] The control $\tilde{u}_\nu(\xi_j)$ in \eqref{eq:sys_eq} is bounded $\forall P_k \in \mathcal{P}$, \ie, $ \sup_{j} \Vert{\tilde{u}_\nu(\xi_\textnormal{d}) - \tilde{u}_\nu(\xi_j)}\Vert  \leq b_\pi, b_\pi \in \mathbb{R} \backslash\{+\infty\}$.\label{a5}
\end{itemize}
The theoretical assumptions are fundamentally grounded in the physical and operational constraints of elastic joint systems. Assumption A1) is strictly satisfied by the collocated motor-side configuration, which yields a well-defined relative degree. The Lipschitz continuity A2) follows from the smooth nature of rigid-link dynamics and passive compliance away from kinematic singularities \cite{pierallini2023iterative}. Reference feasibility A3) is guaranteed by the use of minimum-jerk profiles, while the bounded initialization error A4) captures ubiquitous physical offsets such as sensor noise and gravity-induced sagging. Finally, the bound on RL actions A5) is strictly enforced in software through a $\tanh$ activation mapping the policy output to the physical torque limits, ensuring both theoretical stability and operational safety.
Note that assumptions A1)-A4) are typically used in ILC framework to a single $P_k \in \mathcal{P}$, see, e.g., \cite{ahn1993iterative, pierallini2023iterative}. 
Moreover, thanks to A3), there exist $\forall P_k \in \mathcal{P}$ a desired control input ${u}_\textnormal{d}=\bar{u}_\textnormal{d}+\tilde{u}_\nu(\xi_\textnormal{d})$, such that $\dot{x}_\textnormal{d}=f(x_\textnormal{d})+g(x_\textnormal{d}) [\bar{u}_\textnormal{d} + \tilde{u}_\nu(\xi_\textnormal{d})]$ and $y_\textnormal{d}=h(x_\textnormal{d})$.

\subsection{Optimal control formulation}
Let $J(\xi_j,\nu)$ be a functional cost to be minimized \wrt the control $\tilde{u}_\nu(\cdot)$ that is parametrized by $\nu$, \ie,
\begin{mini}
    {\nu}{J(\xi_j,\nu)}{}{} \label{eq:op}
    \addConstraint{\dot{x}_{j}=f(x_{j})+g(x_{j}) \left[\bar{u}_{j} + \tilde{u}_\nu(\xi_j)\right]}{}{}
    \addConstraint{y_{j}=(x_{j})}{}{}
    \addConstraint{\xi_j =\phi(y_\textnormal{d}, y_{j-I}, \bar{u}_{j-I})}{}{}
    \addConstraint{\forall P_k \in \mathcal{P}}{}{} \,.
\end{mini}
Note that the analytic form of $J(\cdot)$ is not available and has to be approximated. For this reason, the definition of the running cost is more accessible and easier to exploit in an iterative formulation.
The problem can be addressed within an RL framework
to solve a discrete stochastic optimal control problem by exploring and exploiting the sample space, see, \eg, \cite{sutton2018reinforcement}. We can rewrite the functional cost in \eqref{eq:op} as 
$ \hat{J}(\check{\xi}_j, \nu) = \mathop{\mathbb{E}}_{\tilde{u}_\nu(\cdot)} [ c(\check{\xi}_j, {\tilde{u}}_\nu(\cdot)) + \gamma \hat{J}(\check{\xi}_{j+}, \nu) ] $,
where $c(\cdot)$ is the running cost, $\gamma \in [0,1]$ is the discounted factor, and $\tilde{u}_\nu(\cdot) \sim p_{\tilde{u}}(\cdot,\nu) = \mathcal{N}(\cdot\vert(m_\nu(\cdot), \sigma^2_\nu(\cdot)))$
is the gaussian conditional distribution \wrt $\xi_j(\cdot)$. 

\subsection{Goals}
The goal of this work is to design a new learning controller framework capable of:
\begin{itemize}
\label{item:goals}
    \item[G1)] solving the trajectory tracking problem for \eqref{eq:sys_eq}, i.e., finding a control $u(\cdot) : [0,\, t_\textnormal{f}] \to \mathbb{R}^{n}$ such that $\lim_{j\rightarrow \infty} \Vert{\leftindex^k y_\textnormal{d} - \leftindex^k y_j}\Vert_\lambda\leq b_{\textnormal{y}}$, $b_{\textnormal{y}} \in \mathbb{R}\,, k=1,\ldots,m$.
    \item[G2)] reducing the asymptotic convergence limits of the baseline ILC bounds, \ie,
    \begin{align} 
    \label{eq:g_conv_u}\lim _{j \rightarrow+\infty}\norm{\bar{u}_{\textnormal{d}}(t)-\bar{u}_{j}(t)}_{\lambda} & \leq {b}_{\textnormal{u}} < \bar{b}_{\textnormal{u}} \,, \\
    \label{eq:g_conv_x}\lim _{j \rightarrow+\infty}\norm{x_{\textnormal{d}}(t)-x_{j}(t)}_{\lambda} & \leq {b}_{\textnormal{x}} < \bar{b}_{\textnormal{x}} \,, \\
    \label{eq:g_conv_e}\lim _{j \rightarrow+\infty}\textstyle\sum_{i=0}^r\norm{ e^{(i)}_{j}(t)}_{\lambda} & \leq {b}_{\textnormal{e}} < \bar{b}_{\textnormal{e}}\,,
    \end{align}
    where ${b}_{(\cdot)}$ are the new bounds, and $\bar{b}_{(\cdot)}$ represent the asymptotic upper bounds of the standard ILC, see, \eg, \cite{pierallini2023iterative}.
\end{itemize}
Finally, the controller must be mostly feedforward to preserve the robots' elasticity.  

    \section{Problem solution}
\label{sec:p_sol}
In this section, we propose a novel control framework.
We develop the solution step-by-step, \ie, building on the results achieved at each step for the successive ones. The new control input combines model-based and learned terms, \ie, Fig. \ref{fig:general_ctrl_scheme}.
Firstly, we describe how to solve a trajectory tracking control in an iterative process with ILC,  and how, without loss of generality, a-priori knowledge can be included. Secondly, we extend the ILC approach by embedding an RL framework to optimize the initial guess control and the successive contributions. We show that the new control minimizes the upper bound of the convergence for multiple trajectories.

\subsection{Iterative Learning Control framework}
\label{sec:ip}
Recalling system \eqref{eq:sys_eq} under assumption A1)-A5), and a $P_k \in \mathcal{P}$, the pure feedforward control law is
\begin{equation}
\label{eq:eq_ilc}
    \bar{u}_{j+1}(t) = \bar{u}_j(t) + \mathfrak{L}_j(t)\chi_{i}e_j^{(i)}(t)\,,
\end{equation}
where $\mathfrak{L}_j(t) \in \mathbb{R}^{n\times n}$ is the iteration and time varying learning gain, $\chi_{i} \in \mathbb{R}^{n\times n}, \chi_{i} \succ 0, \forall i = 0, \ldots, r$ are the tunable control gains, which affect the convergence velocity \cite{pierallini2023iterative}, and $e_j^{(i)}(t) \in \mathbb{R}^n$ is the error signal, defined as $e_j^{(i)}\doteq y_{\textnormal{d}}^{(i)}-y_{j}^{(i)} \,, \forall i = 0, \ldots, r$. We omit the left index $k$ for the sake of simplicity. The term $\chi_{i}e_j^{(i)}$ follows Einstein summation convention, \ie
\begin{align}
\label{eq:error}
    \chi_{i}e^{(i)}_{j}(t) \doteq& \sum_{i=0}^{r} \chi_{i}e^{(i)}_{j}(t) 
    = \sum_{i=0}^{r} \chi_{i}\left(y_{\textnormal{d}}^{(i)}(t)-y_{j} ^{(i)}(t)\right) \,.
\end{align}

\begin{thm}
\label{thm:conv_ilc}
Consider the system in \eqref{eq:sys_eq} and a task $P_k \in \mathcal{P}$ with $y_\textnormal{d}(t) \in \mathbb{R}^n$ to be the desired output trajectory. Under A1)-A5), if the learning gain $\mathfrak{L}_j(t) \in \mathbb{R}^{n \times n}$ in \eqref{eq:eq_ilc} satisfies
\begin{equation}
\label{eq:ilc_cond}
    \left\|I_{n}-\mathfrak{L}_{\textnormal{j}}(t) \chi_r E(x_{j})\right\| \leq \eta<1, \forall t \in [0,t_\textnormal{f}], \forall j\,,
\end{equation} 
then
\begin{align}
\label{eq:ilc_conv_u}\lim _{j \rightarrow+\infty}\norm{\bar{u}_{\textnormal{d}}(t)-\bar{u}_{j}(t)}_{\lambda} & \leq b_{\textnormal{u}} \in \mathbb{R}_{+} \backslash\{+\infty\}\,, \\
\label{eq:ilc_conv_x}\lim _{j \rightarrow+\infty}\norm{x_{\textnormal{d}}(t)-x_{j}(t)}_{\lambda} & \leq b_{\textnormal{x}} \in \mathbb{R}_{+} \backslash\{+\infty\}\,, \\
\label{eq:ilc_conv_e}\lim _{j \rightarrow+\infty}\textstyle\sum_{i=0}^r\norm{ e^{(i)}_{j}(t)}_{\lambda} & \leq b_{\textnormal{e}} \in \mathbb{R}_{+} \backslash\{+\infty\}\,.
\end{align}
\end{thm}
\begin{proof}
\label{proof:conv_ilc}
Given the controller \eqref{eq:eq_ilc} and \eqref{eq:error} we derive
\begin{align}
    \bar{u}_{j+1} 
    &= \bar{u}_{j} + \mathfrak{L}_j \Big[ \Psi(x_d,x_j) + \chi_r \tilde{E}_j \left(\bar{u}_{j} + \tilde{u}_{\nu}(\xi_j)\right)\Big] \;,  
\end{align}
where $\Psi(x_d,x_j) = \sum_{i=0}^{r} \chi_{i} (L_{f}^{i} h(x_\textnormal{d}) - L_{f}^{i} h(x_{j}) )$ and $\tilde{E}_j = E(x_\textnormal{d}) - E(x_{j})$.
Given A2) and defining $\delta \bar{u}_j \doteq \bar{u}_\textnormal{d} - \bar{u}_j$, $\delta \tilde{u}_j \doteq \tilde{u}_\nu(\xi_\textnormal{d}) - \tilde{u}_\nu(\xi_{j})$, $\delta x_j \doteq x_\textnormal{d} - x_j$, we compute
\begin{align}
\label{eq:ineq_u}
\norm{\delta \bar{u}_{j+1}}
\leq & \; \norm{I_n - \mathfrak{L}_j \chi_r E(x_{j})} \norm{\delta \bar{u}_{j}}  \notag \\
    &+ \norm{\mathfrak{L}_j} \norm{\Psi(x_d,x_j)}  \notag  \\
    &+  \norm{\mathfrak{L}_j} \norm{\chi_r} E_0 \norm{\delta x_{j}} \norm{\bar{u}_\textnormal{d} + \tilde{u}_\nu(\xi_\textnormal{d})} \notag \\
    &+  \norm{\mathfrak{L}_j} \norm{\chi_r} E_0 \norm{\delta x_{j}} \norm{\delta \tilde{u}_j}\,.
\end{align}
Defining 
$\Gamma_j \doteq \textstyle\sup_t\{ (r+1) \Psi^* + \Vert{\chi_r}\Vert E_0 \Vert{\bar{u}_\textnormal{d} + \tilde{u}_\nu(\xi_\textnormal{d})}\Vert\} \,$ and
$ \zeta_j \doteq \Vert{\mathfrak{L}_j \chi_r E(x_{j})}\Vert\,$, 
where $\Vert{\Psi(x_d,x_j)}\Vert  
    \leq (r+1) \Psi^* \Vert{\delta x_{j}}\Vert$ and $\Psi^* = \max_i \Vert{\chi_{i}}\Vert \Vert{\psi_0^i}\Vert$. Recalling \eqref{eq:ilc_cond}, we rewrite \eqref{eq:ineq_u} as
\begin{equation}
\label{eq:ineq_u_compact}
    \norm{\delta \bar{u}_{j+1}} \leq \eta \norm{\delta \bar{u}_{j}} + \Gamma_j \norm{\delta x_j} + \zeta_j \norm{\delta \tilde{u}_j} \,.
\end{equation}
Given assumption A2) we write the following inequality for \eqref{eq:sys_eq}, 
$ \Vert{\delta x_j}\Vert \leq \int_0^{t} (f_0 + g_0 \Vert{\bar{u}_\textnormal{d}(\tau) + \tilde{u}_\nu(\xi_\textnormal{d}(\tau))}\Vert ) \Vert{\delta x_j(\tau)}\Vert + \Vert{g(x_j(\tau))}\Vert ( \Vert{\delta \bar{u}_{j}(\tau)\Vert} + \Vert{\delta \tilde{u}_j(\tau)}\Vert ) \, \textnormal{d}\tau+b_\textnormal{l} $.
Applying Gronwall's Lemma leads to
\begin{align}
\label{eq:gronwall}
    &\norm{\delta x_j}
    \leq b_\textnormal{l}e^{c_2t} 
    \notag\\ 
    & \quad 
    + c_1 \int_{0}^{t} \left(\, \norm{\delta \bar{u}_{j}(\tau)} + \norm{\delta \tilde{u}_j(\tau)} \right) e^{c_2(t-\tau)} \textnormal{d}\tau \,,
\end{align}
where $c_1 \doteq \sup_\tau \Vert{g(x_j)}\Vert \,, c_2 \doteq \sup_{\tau} \left[ f_0 + g_0 \Vert{u_\textnormal{d}}\Vert \right]$.

Substituting \eqref{eq:gronwall} in \eqref{eq:ineq_u_compact} and applying the $\lambda$-norm, we obtain
\begin{align}
\label{eq:ineq_u_lambda}
    &\norm{\delta \bar{u}_{j+1}}_\lambda  
    \leq b_\textnormal{l}\sup_t \{e^{(c_2-\lambda)t}\} +\eta \norm{\delta \bar{u}_{j}}_\lambda + \zeta_j \norm{\delta \tilde{u}_j}_\lambda \notag \\
    & + \Gamma_j c_1 \left(\,\norm{\delta \bar{u}_j}_\lambda + \norm{\delta \tilde{u}_j}_\lambda\right) \sup_t \int_0^{t} 
     e^{(c_2-\lambda)(t-\tau)} \textnormal{d}\tau \,.
\end{align}
Solving integral in \eqref{eq:ineq_u_lambda}, rearranging terms, and assuming that $(c_2-\lambda)\leq0$, leads to
$ \Vert{\delta \bar{u}_{j+1}}\Vert_\lambda \leq  b_\textnormal{l} + \zeta_j \Vert{\delta \tilde{u}_j}\Vert_\lambda  + \eta \Vert{\delta \bar{u}_{j}}\Vert_\lambda + \rho_j(\lambda) (\, \Vert{\delta \bar{u}_j}\Vert_\lambda + \Vert{\delta \tilde{u}_j}\Vert_\lambda) $,
where $C_1(\lambda)\doteq c_1 \sup_t \{(1-e^{-(c_2-\lambda)t})/(\lambda-c_2)\}$, $\rho_j(\lambda) \doteq \Gamma_j C_1(\lambda)$. Defining $\Omega\doteq\sup_j\{\eta+\rho_j(\lambda)\}$, $\Upsilon \doteq \sup_j\{\zeta_j+\rho_j(\lambda)\}$, and recalling A5), we write the following inequalities
\begin{align}
    \norm{\delta \bar{u}_{j+1}}_\lambda 
    &\leq \Omega \norm{\delta \bar{u}_{j}}_\lambda + \Upsilon \norm{\delta \tilde{u}_j}_\lambda +b_\textnormal{l} \\
    \label{eq:ineq_u_final}
    &\leq \Omega \norm{\delta \bar{u}_{j}}_\lambda + \Upsilon b_{\pi} +b_\textnormal{l} \,, 
\end{align}
For hypotheses $\eta < 1$, then $\forall c_2 , \exists \lambda$ large enough such that $\Omega < 1$, and satisfying the assumption $(c_2-\lambda)\leq0$. Substituting in \eqref{eq:ineq_u_final} the previous $j$ trials, we have
\begin{equation}
\label{eq:conv_u_analysis}
    \norm{\delta \bar{u}_{j+1}}_\lambda \leq \Omega^j \norm{\delta \bar{u}_{0}}_\lambda + \dfrac{1-\Omega^{(j-1)}}{1-\Omega}\left(\Upsilon b_{\pi} +b_\textnormal{l}\right) \,,     
\end{equation}
Then, when the index $j$ approaches $+\infty$, we have
\begin{equation}
\label{eq:conv_u}
    \lim_{j\to\infty}\norm{\delta \bar{u}_{j+1}}_\lambda \leq \dfrac{\Upsilon b_{\pi} +b_\textnormal{l}}{1-\Omega} \doteq b_\textnormal{u}\,.
\end{equation}
This proves \eqref{eq:ilc_conv_u}. 

Considering \eqref{eq:conv_u}, we apply $\lambda$-norm and $j\to+\infty$ to \eqref{eq:gronwall}, leads to
\begin{align}
\label{eq:conv_x}
    \lim_{j\to\infty} \norm{\delta x_j}_\lambda
    &\leq b_\textnormal{l} + C_1 (\lambda)
    \left(b_\textnormal{u} + b_\pi\right) \doteq b_\textnormal{x}\,,
\end{align}
This proves \eqref{eq:ilc_conv_x}.

Finally, considering \eqref{eq:conv_u} and \eqref{eq:conv_x}, we apply $\lambda$-norm to $\sum^r_{i=0}e^{(i)}_j$ when $j\to+\infty$, obtaining the following inequality 
\begin{align}
\label{eq:conv_e}
\lim_{j\to+\infty}\textstyle \sum^r_{i=0} \norm{e^{(i)}_j}_\lambda 
&\leq \Delta b_\textnormal{x} + E(x_\textnormal{d}) b_\textnormal{u} \doteq b_\textnormal{e}\,,
\end{align}
where $\Delta \doteq \sum_{i=0}^{r} \psi^i_0 +  E_0 b_\textnormal{u} + E_0 \Vert{u_\textnormal{d}}\Vert.$
This proves \eqref{eq:ilc_conv_e}.

The proof of Theorem \ref{thm:conv_ilc} is completed.
\end{proof}
It is important to note that Theorem 1 holds for any output function $h(x)$ that yields a well-defined relative degree $r$. While the specific value of $r$ and the nature of the zero dynamics influence the synthesis of the learning operators, the fundamental stability structure and the form of the asymptotic error bounds remain invariant \cite{pierallini2025fishing, aarnoudse2024design, axehill2014estimation}.

Given the same assumptions of Theorem \ref{thm:conv_ilc}, we can derive \eqref{eq:ilc_conv_x}, $\forall P_k \in \mathcal{P}$,  \ie,
$ \lim_{j\to\infty} \Vert{\leftindex^k x_{\textnormal{d}} - \leftindex^k x_j}\Vert_\lambda \leq \leftindex^k b_\textnormal{x}$, as shown in Proof \ref{proof:conv_ilc},
where we use $\leftindex^k [\cdot]$ to show explicitly the dependency of $k$-th process.
Recalling A2), one can write the following inequality
\begin{align}
    \lim_{j\rightarrow \infty} \norm{\leftindex^k y_\textnormal{d}(t) - \leftindex^k y_j(t)}_\lambda 
    &\leq \psi_0 \max_k \leftindex^k b_{\textnormal{x}} \doteq b_{\textnormal{y}}\,,
\end{align}
which achieves G1). 
%

\begin{remark}
    Although the convergence results in Theorem~\ref{thm:conv_ilc} align with those of standard ILC, the underlying control problem is fundamentally distinct due to the active presence of the RL contribution $\tilde{u}_{\nu}(\cdot)$ in the loop. Theorem~\ref{thm:conv_ilc} establishes a formal bound on the tracking error, specifically accounting for an additional control action. The only assumption used for the control $\tilde{u}_{\nu}(\cdot)$ is A5). In our implementation, see Sec.~\ref{sec:rilc}, this assumption is strictly enforced by saturating the PPO policy $\tilde{u}_\nu$ via a $\tanh$ activation function. This design guarantees that the RL control action strictly satisfies $\Vert{\tilde{u}_\nu(\xi_\textnormal{d}) - \tilde{u}_\nu(\xi_j)}\Vert \leq b_{\pi}$ even during stochastic exploration. Consequently, while the derived bounds \eqref{eq:ilc_conv_u}-\eqref{eq:ilc_conv_e} are conservative, as shown in \cite{wei2016output, shen2020zero, pierallini2023iterative}, they validate the feasibility of the coupled RL-ILC architecture. 
    Furthermore, the asymptotic error bound $b_{\textnormal{u}}$ in \eqref{eq:conv_u} depends linearly on the bound $b_{\pi}$ via the term $\Upsilon \doteq \sup_j \{ \zeta_j + \rho_j(\lambda) \}$. \label{rem:sensitivity} By expanding the definitions of $\zeta_j$ and $\rho_j(\lambda)$, we obtain that
    given a $\lambda$, it is always possible to tune the control gains $\chi_i$ and the learning gain $\mathfrak{L}_j$ to reduce $\Upsilon$, achieving the same asymptotic convergence bound regardless of the value of $b_{\pi}$.
\end{remark}
%
%

Note that the initial guess $\bar{u}_0(t) \in \mathbb{R}^n$ in law \eqref{eq:eq_ilc} can be arbitrarily chosen. Typically, it relies on partial and nominal knowledge of the model or a constant input required to keep the robot in the desired starting position of the trajectory. It is common for the initial guess to use the available a-priori knowledge, as shown in \cite{pierallini2023iterative}, 
For this reason, the choice of $\bar{u}_0(t) = u_{\textnormal{mb}}(t)$, where $u_{\textnormal{mb}}(t)$ is model-based knowledge, is recommended. In this framework, model-based control can be considered an external contribution \wrt ILC. 
Moreover, a pure-feedforward ILC law cannot reject the initial offset during the current iteration, causing degradation on tracking performance. Instead, non-repetitive initial shift modifies the ILC action of the next iteration, which could have a different initial offset. Conversely, a feedback
term can compensate for the initial shift, improving the overall robustness of the iterative process.
The final control law can then be rewritten as follows
\begin{equation}
\label{eq:total_control}
    u_j(t) = \bar{u}_{j}(t) + \tilde{u}_\nu(\xi_j(t)) + u_{\textnormal{mb}}(t) + u_{\textnormal{fb}}(t)\,,
\end{equation}
where we set $\bar{u}_0(t) = 0$, $\forall P_k \in \mathcal{P}$, and $u_{\textnormal{fb}}(t)$ is the feedback control contribution with low gains to not alter the system compliance drastically.

\subsection{Main result: Reinforced Iterative Learning Control law}
\label{sec:rip}

\begin{algorithm}[ht]
\footnotesize
\SetAlgoLined
\caption{Reinforced Iterative Learning Control}
\label{alg:rilc}
\KwIn{
\\
    \quad \texttt{SYSTEM}\,\,\;
    \quad Controller parameters: $\bar{u}_0$, $\chi_i$, $\mathfrak{L}_j$, $\nu$\,\,\;
    \quad PPO hyperparameters: ($\varepsilon, \gamma, \ldots$)\,\,\;
    \quad Additional parameters: $\delta$, \texttt{max\_iter}, $l$\,\,\;
    \quad Flag \texttt{fl\_rl} \tcp*{for RL updating}
}
\textbf{Initialize:}
\\
    \quad $\mathcal{B}_{\textnormal{ILC}} \gets \emptyset, \,\, \mathcal{B}_{\textnormal{RL}} \gets \emptyset$ \tcp*{buffers}
    \quad $\tilde{u}_{\nu} \gets \nu$ \tcp*{init RL control}
    \quad $I \gets \{0,\ldots,l\}$ \tcp*{set memory dim}
    \quad $\mathcal{P} \gets \{P_1, \dots, P_m\}$ \tcp*{process stack}    
    \quad \texttt{PPO}$\gets (\varepsilon, \gamma, \dots)$ \tcp*{init PPO}
    \While{\{\texttt{len} \{$\mathcal{P}\} > 0 \lor \nu \neq \nu^{*}$\}}{
        \texttt{SYSTEM} $\gets$ \texttt{domain randomization}\,\,\; 
        $y_\textnormal{d} \gets P_k \gets \texttt{PopOutFirst}\{\mathcal{P}\}$\,\,\;
        Reset: $\bar{u}_j \gets \bar{u}_0$, $j \gets 0$, $t \gets 0$, $\phi(\cdot)  \gets y_\textnormal{d}$ \;
            \While{$j\leq$  \textnormal{\texttt{max\_iter}}}{
            \texttt{SYSTEM} $\gets x_0$ \tcp*{set initial condition}
            \While{$t \leq t_{\textnormal{f}}$}{
                $e_j^{(i)} = y_\textnormal{d}^{(i)} - y_j^{(i)}$ \tcp*{error}
                $\xi_j = \phi(y_\textnormal{d}, y_{j-I}, \bar{u}_{j-I})$ \tcp*{Eq. \eqref{eq:sys_eq}} 
                $u_j = \bar{u}_j + \tilde{u}_{\nu}(\xi_j) + u_{\textnormal{mb}} + u_{\textnormal{fb}}$ \tcp*{Eq. \eqref{eq:total_control}} 
                \texttt{SYSTEM} $\gets u_j$ \tcp*{update system}
                $y_j^{(i)} \gets$ \texttt{SYSTEM}\,\,\;
                $c \gets \{\xi_j,\tilde{u}_{\nu}\}$ \tcp*{Eq. \eqref{eq:run_cost}}
                $\phi(\cdot) \gets \{y_j, \bar{u}_j\}$ \tcp*{update $\phi(\cdot)$}
                $\mathcal{B}_{\textnormal{ILC}} \gets (e^{(i)}_j, \bar{u}_j)$ \tcp*{update buffer}
                $\mathcal{B}_{\textnormal{RL}} \gets (c, \xi_j, \tilde{u}_{\nu})$ \tcp*{update buffer}
                \If{\textnormal{\texttt{fl\_rl}}}{$\tilde{u}_{\nu} \gets\nu \gets \texttt{PPO}\{\mathcal{B}_{\textnormal{RL}}, \nu\}$ \tcp*{Eq. \eqref{eq:rl_update}} }
                $t ++$
            }            
            $\bar{u}_{j+1} \gets \{\mathcal{B}_{\textnormal{ILC}}, \chi_i, \mathfrak{L}_j\}$ \tcp*{Eq. ~\eqref{eq:eq_ilc}}
            $j \gets j ++$ 
        }
    }
    \label{algo:RILC}
\end{algorithm}
The most recent successful model-free on-policy RL method is the PPO algorithm \cite{schulman2017proximal, rudin2022learning}, 
due to its proven stability and efficiency in the state-of-the-art. The on-policy nature of PPO is particularly advantageous, as it enables direct leverage of the samples generated by the ILC loop.
Recalling the optimal problem \eqref{eq:op}, we consider the control $\tilde{u}_\nu(\cdot)$ approximated by DNN.
The parameter $\nu$ is updated following the gradient policy methods, \cite{schulman2017proximal}, as follows
\begin{equation}
\label{eq:rl_update}
    \nu \xleftarrow{} \nu - \nabla_\nu \left( \hat{J}^\textnormal{ C} (\check{\xi}_j,\nu_\textnormal{old},\nu)  + L(\nu) \right) \,,
\end{equation}
where $\hat{J}^\textnormal{C}(\cdot,\nu) = $ ${\mathbb{E}}_{\tilde{u}_\nu(\cdot)} [ \max \{ r(\cdot,\nu)\hat{A}(\cdot), $ $
\textnormal{clip} [r(\cdot,\nu)\hat{A}(\cdot), 1 - \varepsilon, 1+\varepsilon ] \}] $,
where $r(\check{\xi}_j,\nu_\textnormal{old},\nu)=p_{\tilde{u}}(\check{\xi}_j, \nu)/p_{\tilde{u}}(\check{\xi}_j, \nu_\textnormal{old})$, $\textnormal{clip}[\cdot]$ is a clipping function, $\varepsilon$ is a hyperparameter, and $\hat{A}(\cdot)=c(\check{\xi}_j,\check{\tilde{u}}_\nu)+\hat{J}(\check{\xi}_{j+},\nu)-\hat{J}(\check{\xi}_j,\nu)$ is the advantage function, \cite{sutton2018reinforcement}.  
$L(\cdot)$ is a regularization loss
to avoid sub-optimal control and improve training stability. 

\begin{remark}
    Recalling \eqref{eq:conv_u_analysis} in Proof \ref{proof:conv_ilc}, we can extend it to define the bounds of $\bar{u}_j(\cdot)$ and $y_j(\cdot)$, $\forall j$. Consequently, all samples $\check{\xi}_{j}(\cdot)$ are limited in a region of the sample space $\forall P_k \in \mathcal{P}$ by the control \eqref{eq:eq_ilc}. This limits the RL exploration and enhances the exploitation of sampled ones.
\end{remark}

\begin{remark}
    In Theorem \ref{thm:conv_ilc}, the update of the parameter $\nu$ does not influence the convergence of the control law \eqref{eq:eq_ilc}.
\end{remark}

Considering the system \eqref{eq:sys_eq}, 
the running cost for a trajectory tracking problem in \eqref{eq:op} is inspired by the theory of NOILC, \cite{amann1996iterative}, as follows
\begin{equation}
\label{eq:run_cost}
    c(\xi_j,\nu) = \sum_{i=0}^r \alpha_i\norm{ e_j^{(i)}(\xi_j,{\nu})} + \beta_j \norm{ \bar{u}_{j}-\bar{u}_{j-1}} \,,
\end{equation}
where we reformulate $e_j^{(i)}(\cdot)$ to explicit the dependence on $(\xi_j,{\nu})$, $\alpha_i$ and $\beta_j$ are positive weights.
In the case of $b_{\textnormal{l}}=0$, the optimal control of \eqref{eq:run_cost} in \eqref{eq:op} has to cancel the dynamics of the system perfectly and follow the desired control, \ie,
\begin{align}
\label{eq:perfect_control}
    \tilde{u}_{\nu^{*}} 
    = g^{\dagger}(x_{j}) \left[ f(x_\textnormal{d}) - f(x_{j}) \right] + {u}_\textnormal{d} - \bar{u}_j \,, \forall t \,,
\end{align}
where ${\nu^{*}}$ is the optimal parameter.
If the \eqref{eq:perfect_control} is satisfied, the error is identically zero $\forall j$, and the ILC is never updated. As a consequence, $\bar{u}_{\textnormal{d}}=0$. If $b_{\textnormal{l}}>0$, the optimal control has to take into account \eqref{eq:perfect_control} and minimize the initial offset over time according to the definition of \eqref{eq:run_cost}. However, \eqref{eq:perfect_control} is practically unusable for multiple reasons. Among these, the desired control is not known a-priori also, and the perfect knowledge of the model is not at our disposal. Further, the exploration and exploitation of sample space through the RL algorithm has to predict internally it to define an approximate function of (sub-)optimal $\tilde{u}_{\nu}(\cdot)$.
\begin{proposition} \label{prop:RILCvsILC}

Consider the system \eqref{eq:sys_eq} under Assumptions A1)--A5). Let ${b}_{\textnormal{u}}$, ${b}_{\textnormal{x}}$, and ${b}_{\textnormal{e}}$ be the asymptotic error bounds defined in \eqref{eq:conv_u}, \eqref{eq:conv_x}, and \eqref{eq:conv_e}, respectively. Let $\bar{b}_{\textnormal{u}}$, $\bar{b}_{\textnormal{x}}$, and $\bar{b}_{\textnormal{e}}$ be the asymptotic error bounds for the standalone ILC, and let ${\Omega}$ and $\bar{\Omega}$ be the contraction rates for the RILC and standalone ILC, respectively. If the following condition holds
\begin{equation}
\label{eq:new_tradeoff_formal}
    (\Upsilon + 1 - \Omega) b_{\pi} < b_{\textnormal{l}} \frac{\bar{\Omega} - {\Omega}}{1-\bar{\Omega}} \,,
\end{equation}
then the RILC framework achieves strictly lower asymptotic error bounds, \ie,
\begin{align}
    {b}_{\textnormal{u}} < \bar{b}_{\textnormal{u}}\,, \label{eq:RILC_u}\\ 
    {b}_{\textnormal{x}} < \bar{b}_{\textnormal{x}}\,, \label{eq:RILC_x}\\ 
    {b}_{\textnormal{e}} < \bar{b}_{\textnormal{e}}\,. \label{eq:RILC_e}
\end{align}

\end{proposition}
\begin{proof}
\label{proof:RILCvsILC}

Setting $b_\pi=0$ in \eqref{eq:conv_u}--\eqref{eq:conv_e} yields the standalone ILC bounds
\begin{align}
    \bar{b}_{\textnormal{u}} &= \frac{b_{\textnormal{l}}}{1-\bar{\Omega}} \,, \label{eq:bar_bu}\\
    \bar{b}_{\textnormal{x}} &= b_{\textnormal{l}} + C_1(\lambda)\bar{b}_{\textnormal{u}} \,, \label{eq:bar_bx}\\
    \bar{b}_{\textnormal{e}} &= \bar{\Delta}\, \bar{b}_{\textnormal{x}} + E(x_{\textnormal{d}})\bar{b}_{\textnormal{u}} \,, \label{eq:bar_be}
\end{align}
where $\bar{\Delta} \doteq \sum_{i=0}^{r} \psi^i_0 + E_0 \bar{b}_{\textnormal{u}} + E_0 \Vert u_{\textnormal{d}} \Vert$.

Given condition \eqref{eq:new_tradeoff_formal}, add $b_{\textnormal{l}}$ to both sides and divide by $1-\Omega$, we have
\begin{equation}
    \frac{(\Upsilon + 1 - \Omega)\,b_{\pi} + b_{\textnormal{l}}}{1-\Omega} < \frac{b_{\textnormal{l}}}{1-\bar{\Omega}} \,,
\end{equation}
from \eqref{eq:conv_u} and \eqref{eq:bar_bu}, the left-hand side equals $b_{\textnormal{u}} + b_\pi$ and the right-hand side equals $\bar{b}_{\textnormal{u}}$, hence
\begin{equation}
    b_{\textnormal{u}} + b_\pi < \bar{b}_{\textnormal{u}} \,, \label{eq:tradeoff_algebra}
\end{equation}
which directly yields \eqref{eq:RILC_u}.

From the definitions of $b_{\textnormal{x}}$ and $\bar{b}_{\textnormal{x}}$ in \eqref{eq:conv_x} and \eqref{eq:bar_bx}:
\begin{equation}
    b_{\textnormal{x}} = b_{\textnormal{l}} + C_1(\lambda)(b_{\textnormal{u}} + b_\pi) < b_{\textnormal{l}} + C_1(\lambda)\bar{b}_{\textnormal{u}} = \bar{b}_{\textnormal{x}} \,,
\end{equation}
where the inequality follows from \eqref{eq:tradeoff_algebra}. This proves \eqref{eq:RILC_x}.

To verify \eqref{eq:RILC_e}, substitute the bounds from \eqref{eq:conv_e} and \eqref{eq:bar_be}:
\begin{equation}
    b_{\textnormal{e}} = \Delta\, b_{\textnormal{x}} + E(x_{\textnormal{d}})b_{\textnormal{u}} \leq
    \bar{\Delta}\,\bar{b}_{\textnormal{x}} + E(x_{\textnormal{d}})\bar{b}_{\textnormal{u}} = \bar{b}_{\textnormal{e}}\,.
\end{equation}
Since $\Delta$ is monotone increasing in $b_{\textnormal{u}}$ and $b_{\textnormal{u}} < \bar{b}_{\textnormal{u}}$ from \eqref{eq:RILC_u}, we have $\Delta < \bar{\Delta}$. Together with $b_{\textnormal{x}} < \bar{b}_{\textnormal{x}}$ from \eqref{eq:RILC_x} and $b_{\textnormal{u}} < \bar{b}_{\textnormal{u}}$, it follows that $b_{\textnormal{e}} < \bar{b}_{\textnormal{e}}$, proving \eqref{eq:RILC_e}.
This completes the proof.

\end{proof}
Proposition \ref{prop:RILCvsILC} achieves G2), showing that RILC yields tighter asymptotic error bounds than standalone ILC.
Finally, a domain randomization can be employed to facilitate the sim-to-real transfer, ensuring that the learned RL control is robust and effective when applied to real-world scenarios. This control architecture is summarized in the pseudocode of Algorithm~\ref{alg:rilc}. The RL policy receives the previous and actual ILC trial information to compute a real-time correction, creating a coupling between the two learning loops.

    \section{Validation} 
\label{sec:val}

\begin{figure*}[t]
\centering
\subfloat[\label{fig:scheme_training}]{\includegraphics[width=0.35\textwidth ]{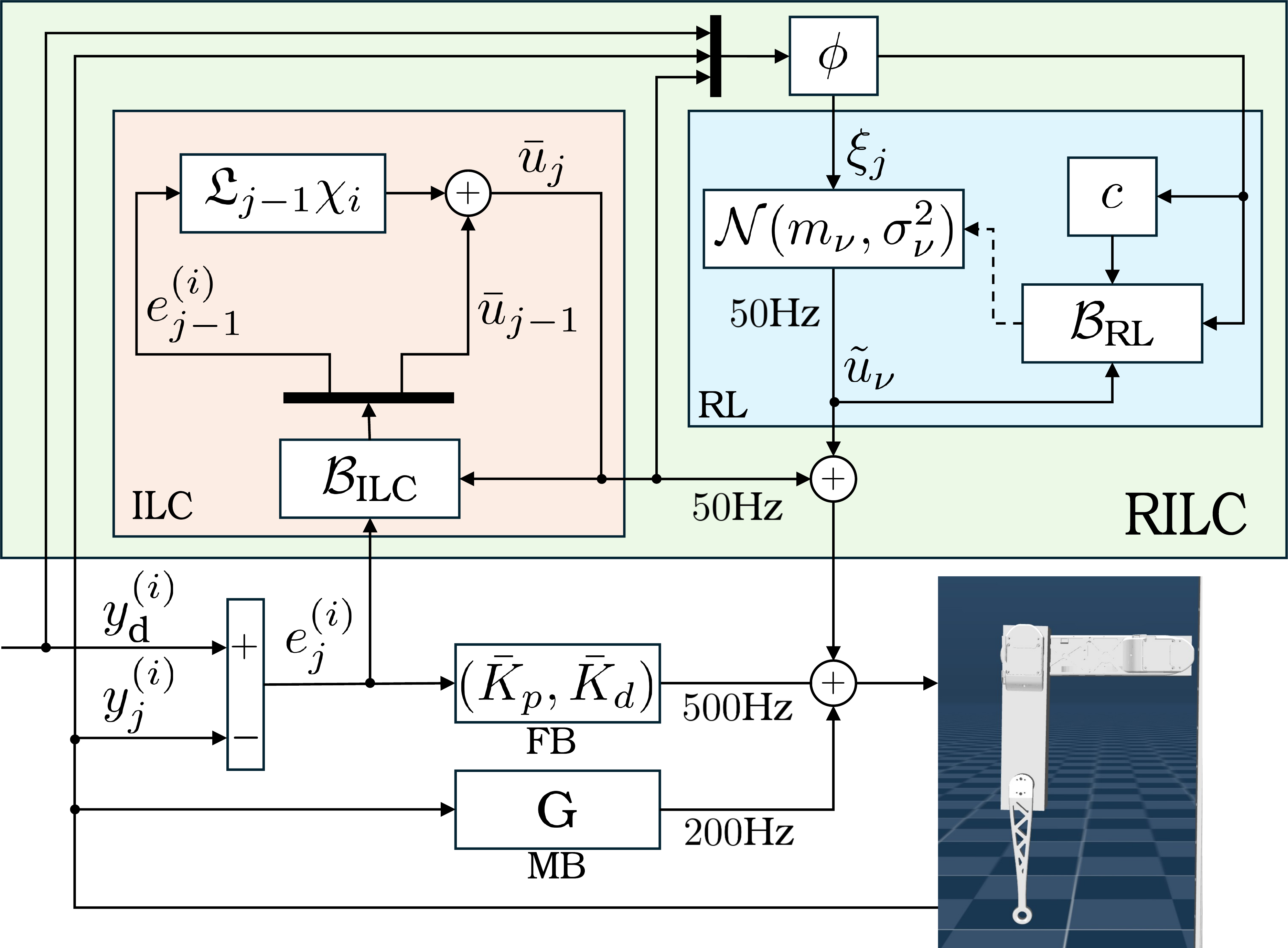}}
\qquad
\subfloat[\label{fig:scheme_experiments}]{\includegraphics[width=0.35\textwidth ]{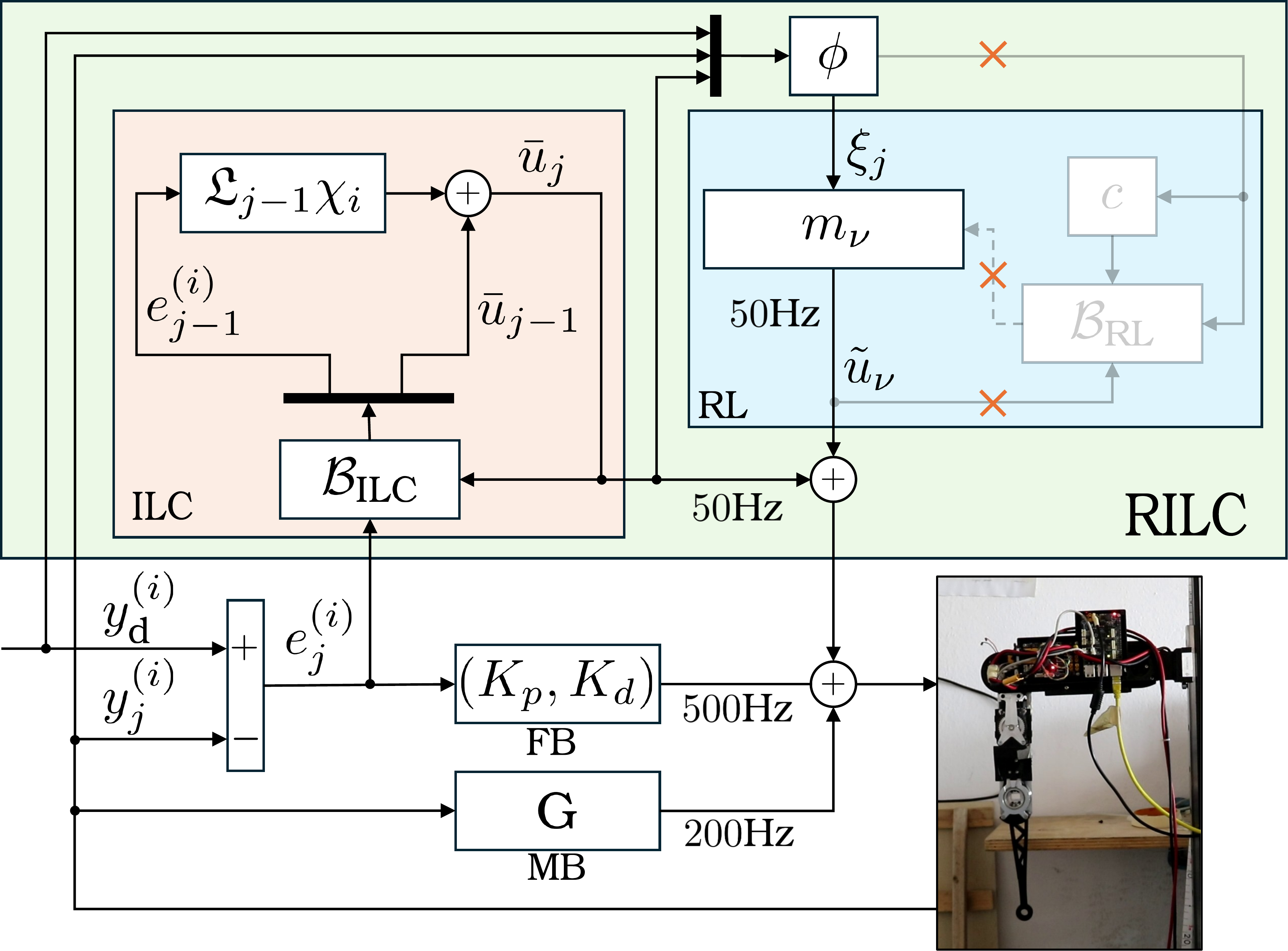}}
\caption{
Fig.s~\ref{fig:scheme_training} and~\ref{fig:scheme_experiments} represent the control scheme of Reinforced Iterative Learning Control (RILC) used in the training phase and experiments, respectively. In simulations, a stochastic policy (to explore the sample space) and parameters from Tab.~\ref{tab:param_test} are employed, while a deterministic policy is deployed in real-world experiments. For further details, see Alg.~\ref{algo:RILC}.}
\label{fig:ctrl_scheme}
\end{figure*}

In this section, we validate the effectiveness of the RILC by comparing it with baseline ILC and RL. 
Moreover, only for numerical experiments, we implement NOILC to highlight the improvements achieved by the proposed learning-based approach. We start by providing a detailed description of the setup used and its subsequent employment in real-world experiments. The metric used to evaluate tracking performance is root mean square error (RMSE).

\subsection{Setup}
\label{sec:setup} The experiments are conducted on a 4-DoFs arm, Fig. \ref{fig:ctrl_scheme}, with two EM-Acts \cite{martignetti2024act}, a modular lightweight SEA. Note that this robot is underactuated; we can set the motor position, which is connected to link one via a linear spring~\cite{martignetti2024act}. However, the actuator motor dynamics are not exactly known, introducing a mismatch.
The experimental platform relies exclusively on direct measurements of the motor-side position, acquired using an AS5047P absolute magnetic on-axis encoder integrated into the mjbots moteus r4.4 controllers. This sensor provides a 14-bit resolution. Motor velocity is estimated through the controller's onboard second-order digital PLL filter, providing a clean feedback signal for the low-level control loop. No link-side sensors or external state observers are employed in the control scheme.

The desired trajectory is a minimum jerk signal in motor space, $y(t)=\theta(t)$, \label{rem:motor_side_choice} that starts from $\theta(0) = 0_{2 \times 1}$ and reaches the final position $\theta _{\textnormal{d}}(t_{\textnormal{f}})$ in $t_{\textnormal{f}}= 1$\si{\second}, \ie, $\theta_{\textnormal{d}}(t) = \theta_{\textnormal{d}}(t_{\textnormal{f}}) [10 ({t}/{t_{\textnormal{f}}})^{3}- 15 ({t}/{t_{\textnormal{f}}})^{4}+ 6 ({t}/{t_{\textnormal{f}}}
)^{5}]$. This collocated configuration is suited for applications where the task tolerance is compatible with the joint deflection. Motor-side tracking avoids the need for state observers, providing a clean experimental baseline for the RL-ILC architecture. While link-side control is theoretically possible, as demonstrated in Appendix~\ref{sec:sim_link}, it would require state observers and introduce sensitivity to estimation errors, making it difficult to isolate the learning performance.
The final position is randomly selected during policy training and by a human operator in real-world tests. A minimum-jerk tracking task allows for a quantification of the synergistic contribution between ILC and RL.
It is important to note that the execution time imposes a dynamic challenge for a compliant, under-actuated manipulator. Furthermore, the system is underactuated and can have a shift in the initial position of $q(0)$. 
While the 1-second minimum jerk trajectories employed in our tests possess a primary frequency content concentrated below the natural frequency of the system, the joint compliance dynamics remain present. Even at low velocities, the nonlinear link dynamics interact with the elastic transmission and reflect onto the motor side as state-dependent load disturbances governed by the spring coupling.
On the control side, the model-based (MB) control in \eqref{eq:total_control} is defined by $u_{\textnormal{mb}}(t) = G(t)$, \ie, gravity compensation, the updating law of ILC \eqref{eq:eq_ilc} is chosen as PD-type, $\bar{u}_{j}(t) = \bar{u}_{{j-1}} (t) + \chi_{0}e_{j-1}(t)+ \chi_{1}\dot{e}_{j-1}(t)$, where $\chi_{0}= 0.075 I_{2}$ and $\chi_{1}= 0.2 I_{2}$ are time-and-episode-invariant diagonal matrix, while higher-order gains are set to zero. We omit the matrix $\mathfrak{L}=I_{2}$ for the sake of clarity. In addition, $\bar{u}_{0}(t)$ is chosen to be null, and the starting configuration is an equilibrium point of the system. We also added a PD feedback control (FB), \ie, $u_{\textnormal{fb}}(t) = K_{\textnormal{p}}e(t) + K_{\textnormal{d}} \dot{e}(t)$, defined by a constant diagonal matrix $K_{\textnormal{p}}= 0_{2 \times 2}$ and $K_{\textnormal{d}}= 10^{-2}\textnormal{diag}([2.5,\, 2.5])$. In real-world tests, the RL control is defined by a deterministic policy, \ie, $\tilde{u}_{\nu}(\xi_{j}) = m_{\nu}(\xi_{j})$. We consider the feedback control as low-level; RL and ILC, which are updated at a lower frequency, are considered high-level control, see Tab.~\ref{tab:param_test} and Fig.~\ref{fig:ctrl_scheme}.

\begin{table}
    \renewcommand{\arraystretch}{1.}
    \vspace{0.5em}
    \centering
    \caption[]{Parameters for Reinforced Iterative Learning Control (RILC) training with PPO algorithm. We denote with $\mathcal{U}
    (a, b)$ the uniform distribution with the bound $a$ and $b$.}
    \footnotesize
    \begin{tabular}{lc}
        \toprule \textbf{RL Parameters}                                                                            & \textbf{Value}                             \\
        \midrule                                                                                                    
        low level control frequency [\si{\hertz}]                                                                  & $500$                                      \\
        high level control frequency [\si{\hertz}]                                                                 & $200$                                      \\
        policy frequency [\si{\hertz}]                                                                             & $50$                                       \\
        [3pt] RL control limits [\si{\newton}]                                                                     & $[-2, 2]$                                  \\
        actuator position limits [\si{\radian}]                                                                  & $[-\pi,\pi]$                               \\
        [3pt]                                                                                                       
        number of policy update                                                                                    & $200$                                      \\
        number of environments                                                                                     & $64$                                       \\
        number of episodes with fixed task                                                                         & $5$                                        \\
        number of episodes with fixed policy                                                                       & $5$                                        \\
        [3pt]                                                                                                       
        policy model                                                                                               & MLP                                        \\
        hidden layer for actor neural network                                                                      & $[64, 64]$                                 \\
        hidden layer for critic neural network                                                                     & $[64, 64]$                                 \\[3pt]
        joint friction [\si{\newton\meter}]                                                                        & $10^{-3}(1 \pm \mathcal{U}(0,0.2))$        \\
        {$\bar{K}_{\textnormal{p}}$} gain [\si{\newton\meter\per\radian}]                                          & $K_{\textnormal{p}}(1 + \mathcal{U}(0,1))$ \\
        $\bar{K}_{\textnormal{d}}$ gain [\si{\newton\meter\second\per\radian}]                                     & $K_{\textnormal{d}}(1 + \mathcal{U}(0,1))$ \\
        \bottomrule                                                                                                 
        \multicolumn{2}{l}{\footnotesize{For more implementation details see \href{https://github.com/yursds/RILC_training}{Github}.}} \\
    \end{tabular}
    \label{tab:param_test}
\end{table}

\subsection{Training}
\label{sec:rl_setup}

\begin{figure}
    \centering
    \includegraphics[width=0.8\linewidth]{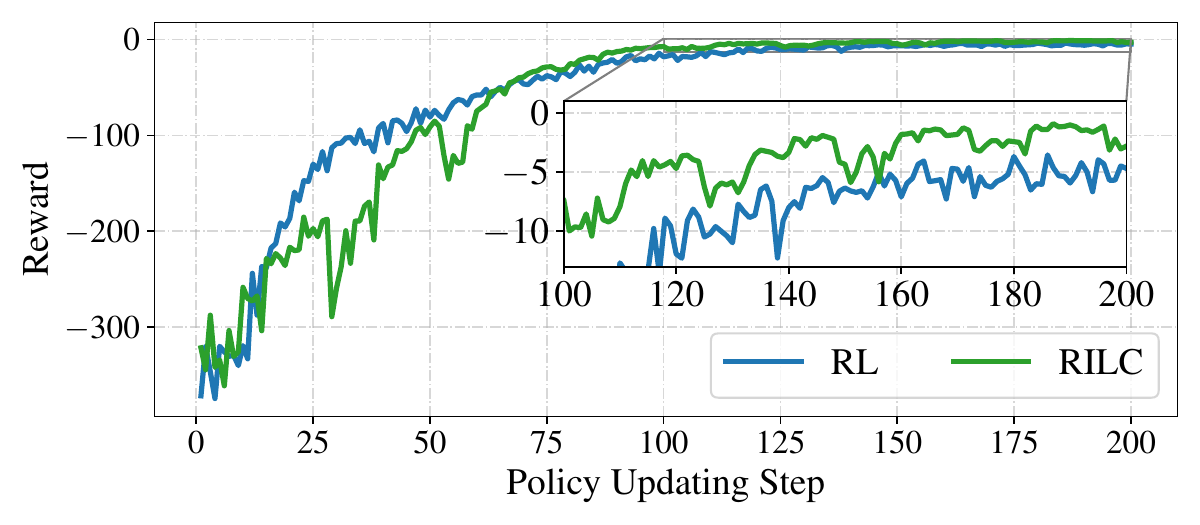}
    \includegraphics[width=0.8\linewidth]{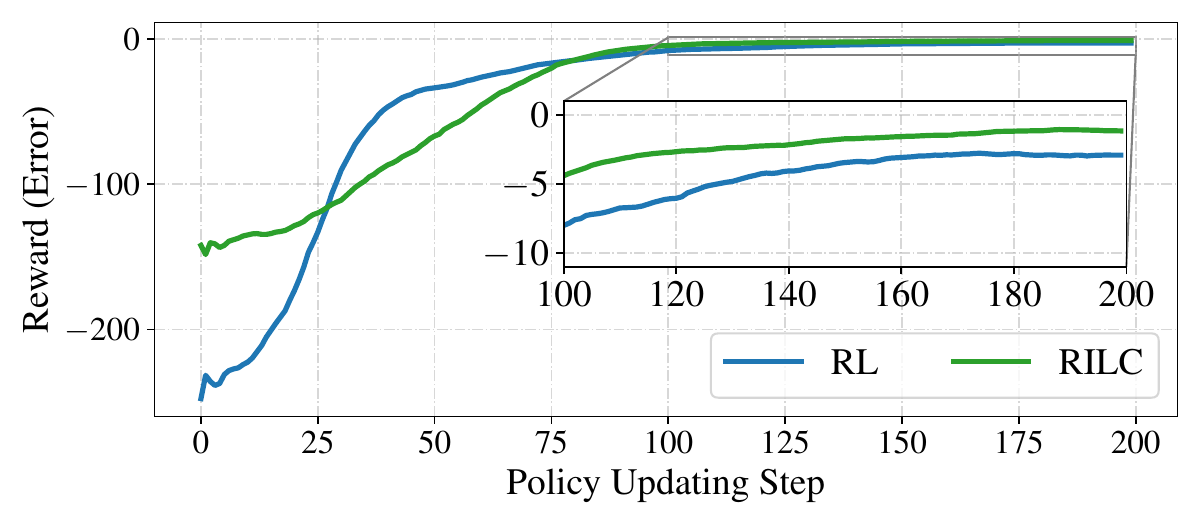}
    \caption{Reward functions evolution of baseline Reinforcement Learning (RL) and Reinforced Iterative Learning Control (RILC) approaches \wrt policy updating. The total 'Reward', \eqref{eq:val_e}-\eqref{eq:sp_e}, and partial cumulative reward dependent on the error, i.e., 'Reward (Error)', \eqref{eq:val_e}, are computed as the mean across all training environments during the simulation.}
    \label{fig:tb_err}
\end{figure}

The training phase of the RILC control is done using the PPO algorithm and
the dynamics of the robot is simulated
using the available data of actuator identification in \cite{martignetti2024act}. The parameters used during training are summarized in Table~\ref{tab:param_test}; for other values, see the default settings in \cite{stable-baselines3}. During the training, we applied domain randomization to ensure the sim-to-real transfer by perturbing joint friction values and adjusting PD feedback gains. For further details, please refer to Tab.~\ref{tab:param_test}.

We define the function
$\xi_{j}(t)=\phi(y_{\textnormal{d}}, y_{j-I}, \bar{u}_{j-I})$, with $I=\{0,1\}$,
\ie
$\xi_{j}(t)= (y_{\textnormal{d}}, y_{j}\,, e_{j-1}\,, \bar{u}_{j}\,, \bar{u}_{j-1}
) \,$.
Note that if $I=\{0\}$, the RL control does not consider the evolution of ILC \wrt iterations. As a result, it may fail to obtain a constructive collaboration between RL and ILC. Hence, the ILC contribution could be seen as an external disturbance. Furthermore, to ensure coherence between the current RL control and $\phi(\cdot)$ inputs of past iterations, the parameter $\nu$ has to remain unchanged for the current iterative process, or a mechanism for $\phi(\cdot)$ is needed to take into account policy updating. 
The control policy is linearly interpolated to reach the low-level control frequency. To mitigate interpolation delay and ensure effective feedforward control, the policy is trained on discre-time shift observations based on pre-computed evolution of the ILC and the desired trajectory.

The running cost used is divided into dense and sparse types, \eg in \cite{rudin2022learning}: running cost obtained in each time step $t$, \ie, continuously, and running cost provided only upon reaching a specific goal or flag, \ie, event-driven. In Sec. \ref{sec:p_sol}, the explicit running cost introduced are extended as follows:
\begin{align}
    \label{eq:val_e}c_{\textnormal{e}}(t) & = 6.0 \norm{e_j(t)}_{2}^{2}+ 0.1 \norm{\dot{e}_j(t)}_{2}^{2}\,, \\
    \label{eq:val_uilc}c_{\bar{u}}(t)     & = 0.1 \norm{\bar{u}_{j}(t)-\bar{u}_{{j-1}}(t)}_{2}^{2}\,,       \\
    \label{eq:val_url}c_{\tilde{u}}(t)    & = 0.1 \norm{\dot{\tilde{u}}_{\nu}(t)}_{2}^{2}\,,
\end{align}
where $c(t)=c_{\textnormal{e}}(t)+c_{\tilde{u}}(t)+c_{\bar{u}}(t)$. The weights in the reward function \eqref{eq:val_e}-\eqref{eq:val_uilc} were empirically tuned to balance tracking accuracy and control smoothness.
We define a sparse running cost for $t=0$, to discourage RL control usage from the start equilibrium configuration, and $t=t_{\textnormal{f}}$, to encourage reducing the root mean square error (RMSE) over time and to minimize the final position error, as follows:
\begin{align}
    \label{eq:sp_e1}c_{\textnormal{sp}_1}(0)                 & = \norm{{\tilde{u}_{\nu}}(0)}_{2}^{2}\,,                                                                     \\
    \label{eq:sp_e}c_{\textnormal{sp}_2}(t_{\textnormal{f}}) & = 10 \left[\left(\textnormal{RMSE}\{e_{j}(t)\} + 1\right)^{-1}+ \norm{e_j(t_\textnormal{f})}_{2}^{2}\right] \,.
\end{align}

\subsection{Simulation results}
\label{sec:sim_res}

\begin{table*}[t]
    \renewcommand{\arraystretch}{1.}
    \setlength{\tabcolsep}{4pt}
    \centering
    \caption{RMSE statistics for Iterative Learning Control (ILC), Norm-Optimal Iterative Learning Control (NOILC), Combined Iterative Learning Control (C-ILC), Reinforcement Learning (RL), and Reinforced Iterative Learning Control (RILC), computed over 3000 trajectories. Suffixes '-N' and '-M' denote nominal and mismatched conditions, while numbers indicate the iteration index.}
    \label{tab:resum_train}
    \footnotesize
    \subfloat[Nominal \label{tab:testN}]{
    \begin{tabular}{lcccccc}
        \toprule \textbf{Test-N} & \textbf{Median} & \textbf{Mean} & \textbf{Std} & \textbf{Min} & \textbf{Max} \\
        \midrule
        ILC-N0 & 0.798 & 0.773 & 0.284 & 0.026 & 1.519 \\
        ILC-N2 & 0.251 & 0.269 & 0.097 & 0.017 & 0.476 \\
        ILC-N20 & 0.003 & 0.004 & 0.003 & 0.001 & 0.017 \\
        [3pt]
        NOILC-N0 & 0.722 & 0.709 & 0.282 & 0.022 & 1.468 \\
        NOILC-N2 & 0.234 & 0.244 & 0.098 & 0.012 & 0.453 \\
        NOILC-N20 & \textbf{0.002} & \textbf{0.002} & \textbf{0.001} & \textbf{0.0001} & \textbf{0.008} \\
        [3pt]
        C-ILC-N0 & 0.586 & 0.596 & 0.289 & 0.076 & 1.233 \\
        C-ILC-N2 & 0.210 & 0.213 & 0.089 & 0.035 & 0.513 \\
        C-ILC-N20 & 0.059 & 0.064 & 0.037 & 0.005 & 0.185 \\
        [3pt]
        RL-N0 & 0.251 & 0.279 & 0.211 & \textbf{0.005} & 1.050 \\
        RL-N2 & 0.251 & 0.279 & 0.211 & \textbf{0.005} & 1.050 \\
        RL-N20 & 0.251 & 0.279 & 0.211 & 0.005 & 1.050 \\
        [3pt]
        RILC-N0 & \textbf{0.182} & \textbf{0.218} & \textbf{0.162} & 0.014 & \textbf{0.713} \\
        RILC-N2 & \textbf{0.127} & \textbf{0.137} & \textbf{0.093} & 0.010 & \textbf{0.399} \\
        RILC-N20 & 0.018 & 0.023 & 0.020 & 0.001 & 0.079 \\
        \bottomrule
    \end{tabular}
    }
    \qquad
    \subfloat[Mismatched \label{tab:testM}]{
    \begin{tabular}{lcccccc}
        \toprule \textbf{Test-M} & \textbf{Median} & \textbf{Mean} & \textbf{Std} & \textbf{Min} & \textbf{Max} \\
        \midrule
        ILC-M0 & 0.797 & 0.773 & 0.284 & 0.046 & 1.511 \\
        ILC-M2 & 0.268 & 0.279 & 0.101 & 0.031 & 0.600 \\
        ILC-M20 & 0.089 & 0.112 & 0.068 & 0.027 & 0.551 \\
        [3pt]
        NOILC-M0 & 0.723 & 0.710 & 0.281 & 0.044 & 1.459 \\
        NOILC-M2 & 0.249 & 0.256 & 0.102 & 0.028 & 0.574 \\
        NOILC-M20 & 0.086 & 0.107 & 0.064 & 0.024 & 0.515 \\
        [3pt]
        C-ILC-M0 & 0.586 & 0.596 & 0.288 & 0.070 & 1.230 \\
        C-ILC-M2 & 0.213 & 0.216 & 0.090 & 0.031 & 0.561 \\
        C-ILC-M20 & 0.072 & 0.082 & 0.044 & \textbf{0.018} & 0.334 \\
        [3pt]
        RL-M0 & 0.256 & 0.287 & 0.207 & \textbf{0.017} & 1.059 \\
        RL-M2 & 0.256 & 0.287 & 0.206 & 0.020 & 1.042 \\
        RL-M20 & 0.256 & 0.287 & 0.207 & 0.020 & 1.055 \\
        [3pt]
        RILC-M0 & \textbf{0.188} & \textbf{0.225} & \textbf{0.159} & 0.022 & \textbf{0.756} \\
        RILC-M2 & \textbf{0.137} & \textbf{0.152} & \textbf{0.093} & \textbf{0.016} & \textbf{0.449} \\
        RILC-M20 & \textbf{0.063} & \textbf{0.071} & \textbf{0.035} & \textbf{0.018} & \textbf{0.311} \\
        \bottomrule
    \end{tabular}
    }
\end{table*}

In this subsection, we benchmark the proposed RILC against RL, ILC, NOILC, and C-ILC \cite{tsurumoto2024combined} as a multi-reference baseline. Implementation details of all baseline controllers are provided in Appendix~\ref{app:baselines}.

We test the performance of all controllers in the whole workspace, evaluating a set of $3000$ randomized trajectories. The training of the C-ILC, RL and RILC is restricted in a reduced workspace, \ie, $y_{\textnormal{d}}\in \{(y_{1}, y_{2}): \Vert{y_1-\pi/2 }\Vert\leq \pi/3 \text{ and}\Vert {y_2}\Vert\leq \pi/3 \}$. We reduce the number of environments to $8$ to ensure fair exploration of the workspace size.
To validate the robustness of the proposed controller, we tested it in the presence of model mismatches, which were not considered in the training. We set a gaussian disturbance to the control input, \ie, $d(t) \sim \mathcal{N}(0,0.5)$~[\si{Nm}]. 
This disturbance is used to simulate the unstructured uncertainties inherent in SEA, \eg, unmodeled friction.

\begin{figure*}[t]
\centering
\subfloat[\label{fig:color_map}]{\includegraphics[width=0.52\textwidth, valign=c]{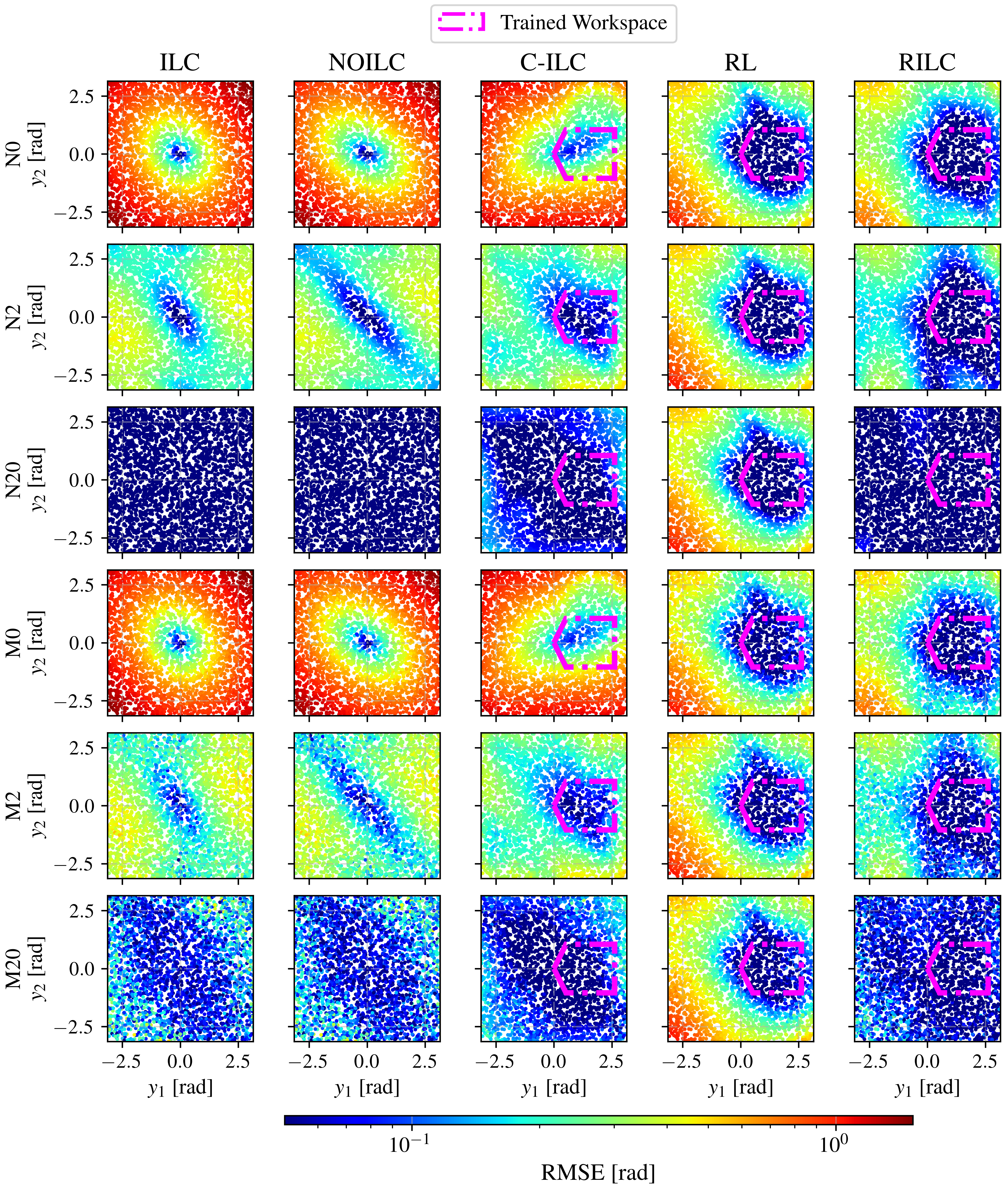}}
\begin{tabular}{c}
    \subfloat[\label{fig:track_sim}]{
        \includegraphics[width=0.42\textwidth]{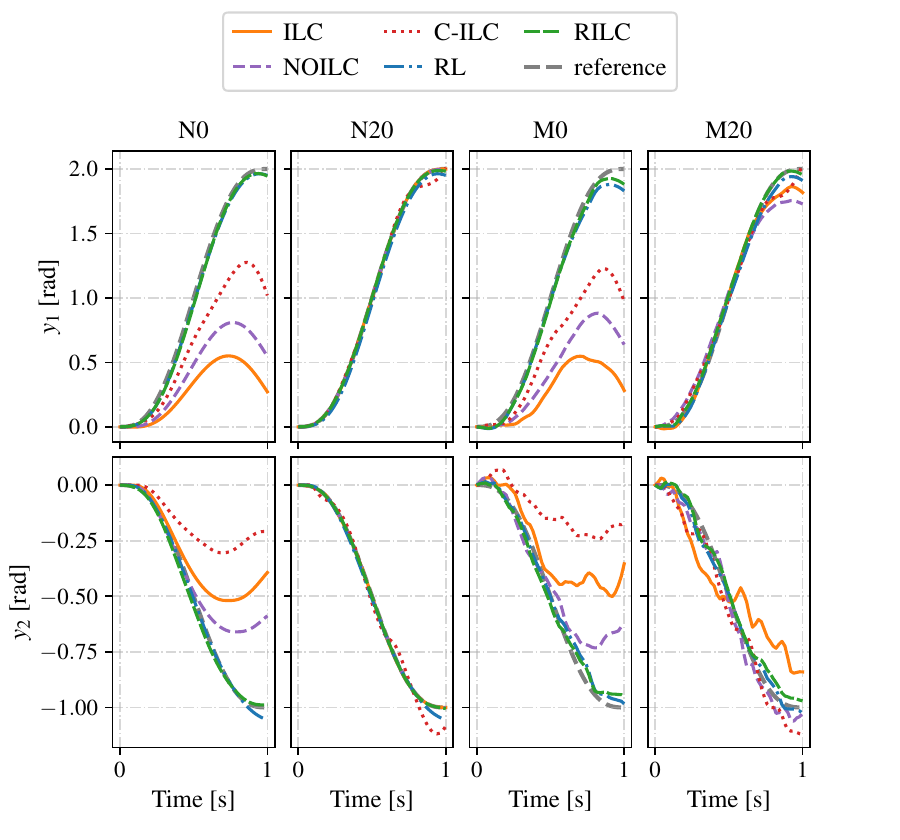}
    }\\
    \vspace{0.2cm} 
    \subfloat[\label{fig:rmse_comparison_sim}]{
        \includegraphics[width=0.4\textwidth]{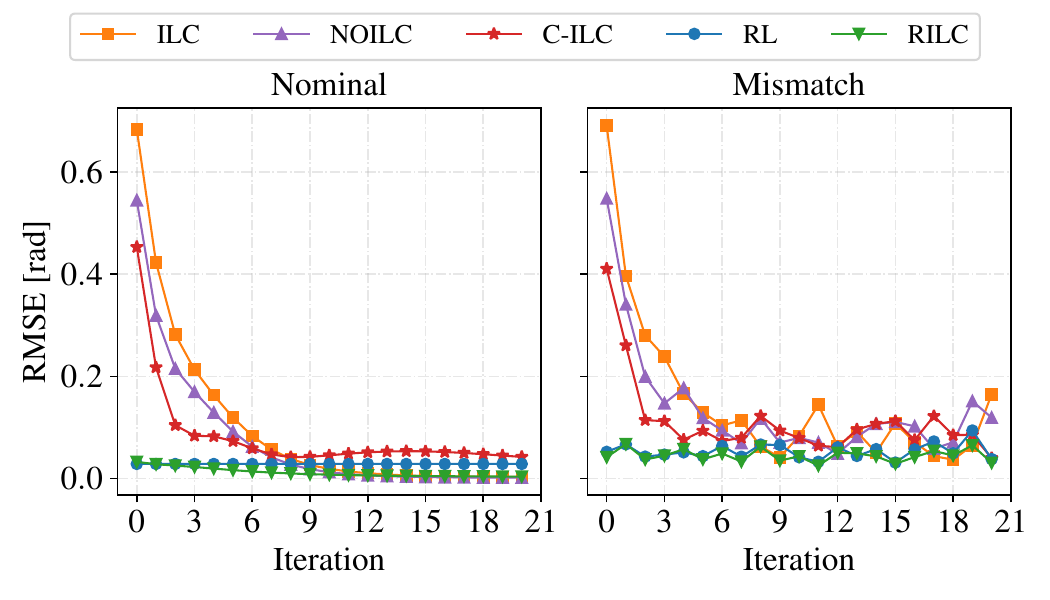}
    }
\end{tabular}

\caption{Simulation results analysis for different controllers: Iteative Learning Control (ILC), Norm-Optimal Iterative Learning Control (NOILC), Combined Iterative Learning Control (C-ILC), Reinforcement Learning (RL), Reinforced Iterative Learning Control (RILC). 
(a) RMSE comparison over 3000 randomized trajectories: each point indicates a $y_{\textnormal{d}}(t_{\textnormal{f}})$, color-mapped by the resulting RMSE. 
(b)-(c) Tracking performance and RMSE evolution for the specific trajectory $y_{\textnormal{d}}(t_{\textnormal{f}}) = [2.0, -1.0]^\top$. 
The suffixes 'N' and 'M' denote nominal and mismatched conditions, while the numerical suffixes indicate the iteration index.}
\end{figure*}

The statistical results in terms of terminal desired position RMSE, (median (Median), mean (Mean), standard deviation (Std), minimum (Min), maximum (Max)), are summarized in Tab.~\ref{tab:resum_train}. The test cases are categorized by suffixes to denote the experimental condition, 'N' and 'M' to indicate nominal and mismatched conditions, and the iteration number.
Fig.~\ref{fig:color_map} visually compares these results in terms of RMSE for the corresponding terminal desired positions.
Specifically, for the trajectory $y_{\textnormal{d}}(t_{\textnormal{f}}) = [2.0, -1.0]^\top$, Fig.s~\ref{fig:track_sim} and \ref{fig:rmse_comparison_sim} illustrate the trajectory tracking performance and the RMSE evolution over iterations, respectively, comparing the response in both nominal and mismatched scenarios.

\subsection{Experiment results}
\label{sec:exp_res}
In this subsection, we report the results of the real-world employment of RILC, \ie,
Alg. \ref{algo:RILC}, using the setup described in Sec. \ref{sec:setup}. We
compare the results for different controllers, i.e., RL, ILC, and RILC.
We report two types of tests, namely Test 1 and Test 2, each of which validates
a specific feature of the proposed framework.

\begin{figure}[t]
\centering
\includegraphics[width=0.45\textwidth]{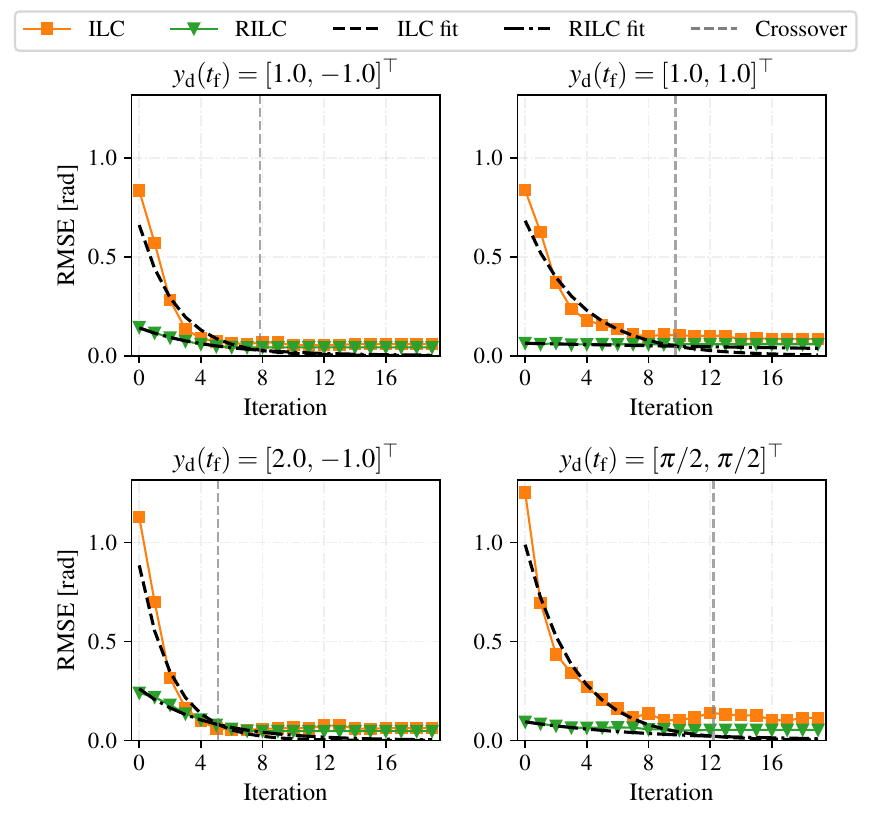}
\caption{RMSE and crossover analysis for the four real-robot target configurations. Dashed and dash-dotted lines show the fitted first-order models $\varepsilon_j = \varepsilon_0 \Omega^{j}$ for Iterative Learning Control (ILC) and Reinforced Iterative Learning Control (RILC). The vertical line marks the crossover iteration $j^*$ of the fitted models.}
\label{fig:crossover}
\end{figure}

\begin{figure*}
    \centering
    \subfloat[\label{fig:contrib_torque}]{
    \includegraphics[width=0.635\linewidth]{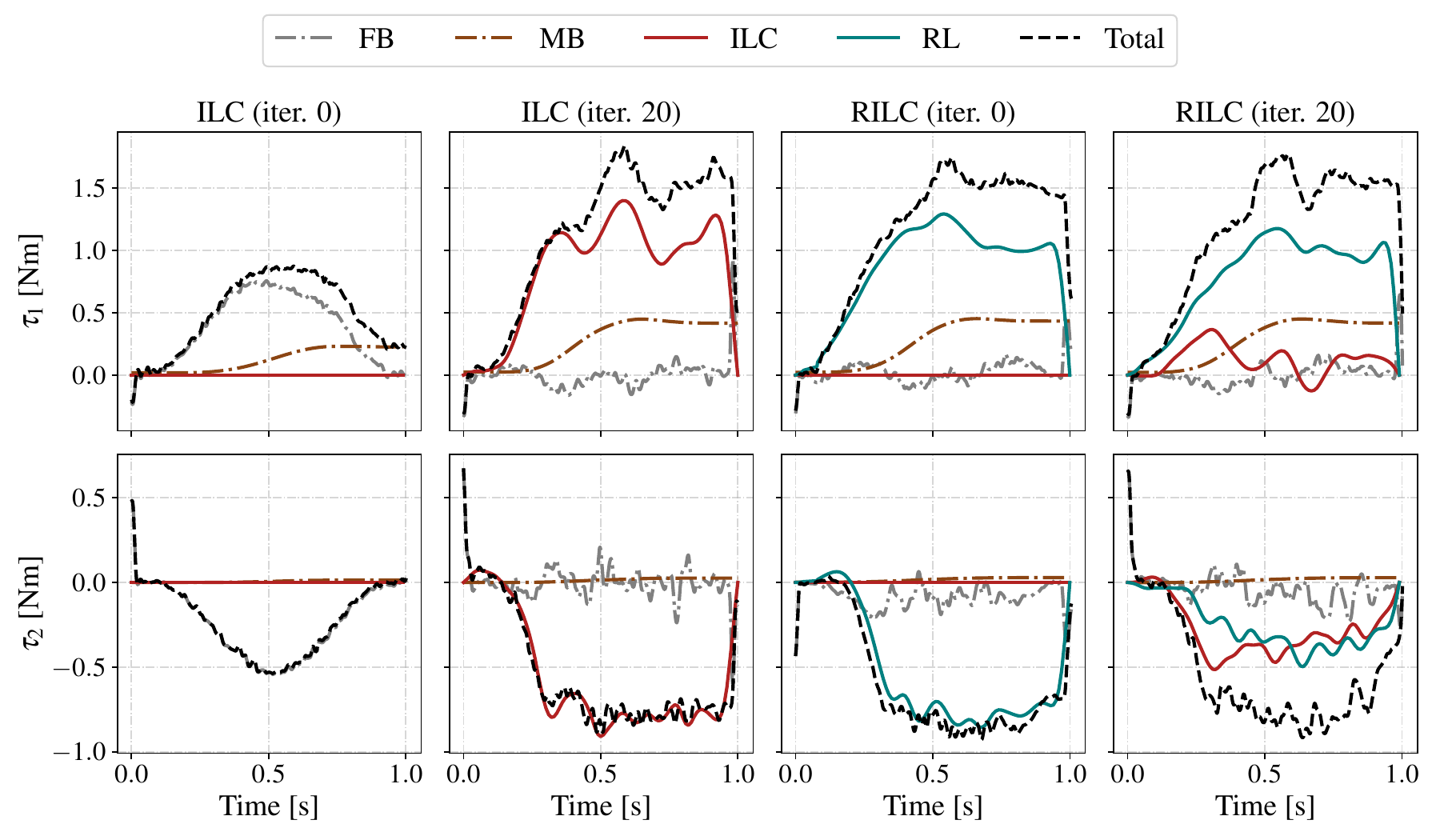}
    }
    \subfloat[\label{fig:track_}]{
    \includegraphics[width=.32\linewidth]{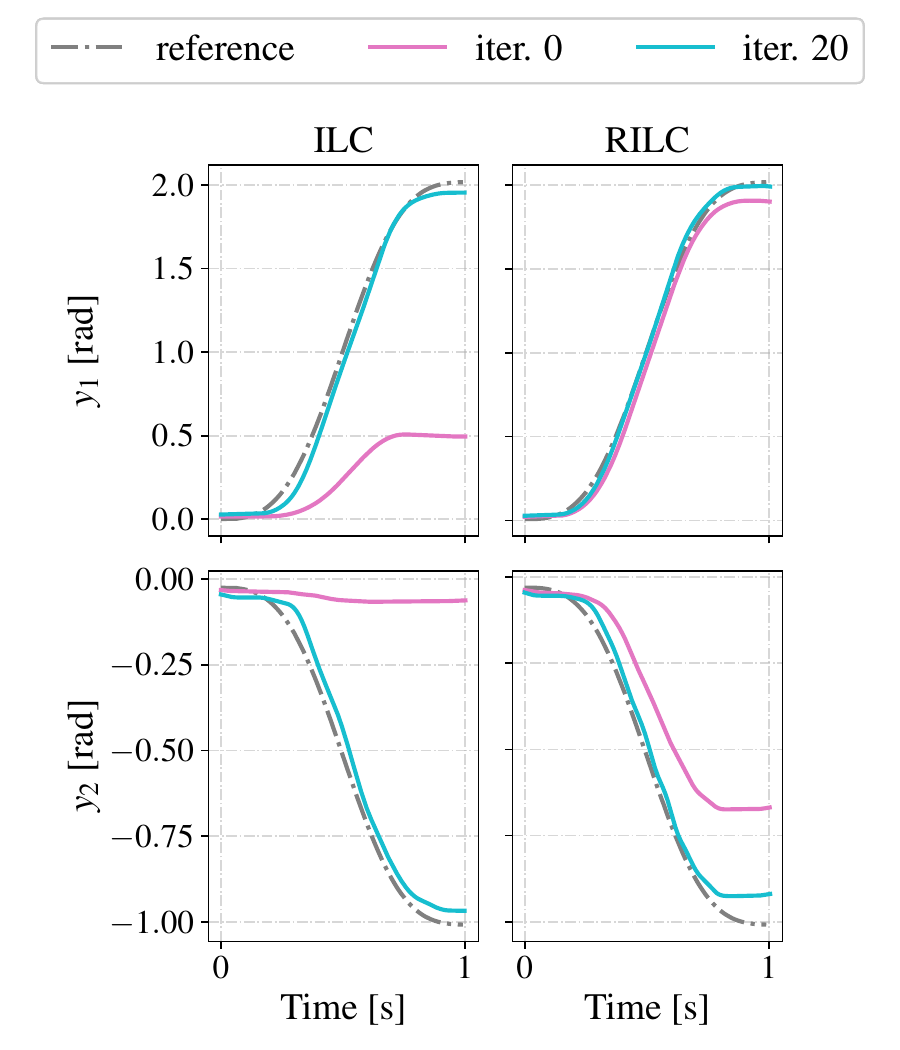}
    }
    \caption{Experimental analysis for the tracking task with target $y_{\textnormal{d}}(t_{\textnormal{f}})=[2.0,-1.0]^{\top}$. (a) Decomposition of control contributions: Feedback (FB), Model-Based (MB), Iterative Learning Control (ILC), Reinforcement Learning (RL), and Total control, at iterations $0$-th and $20$-th. 
    (b) Output tracking evolution comparing the baseline ILC and the proposed RILC approach. 
    \label{fig:control_and_tracking_analysis}
    }
\end{figure*}

\begin{figure}
    \includegraphics[width=0.24\linewidth]{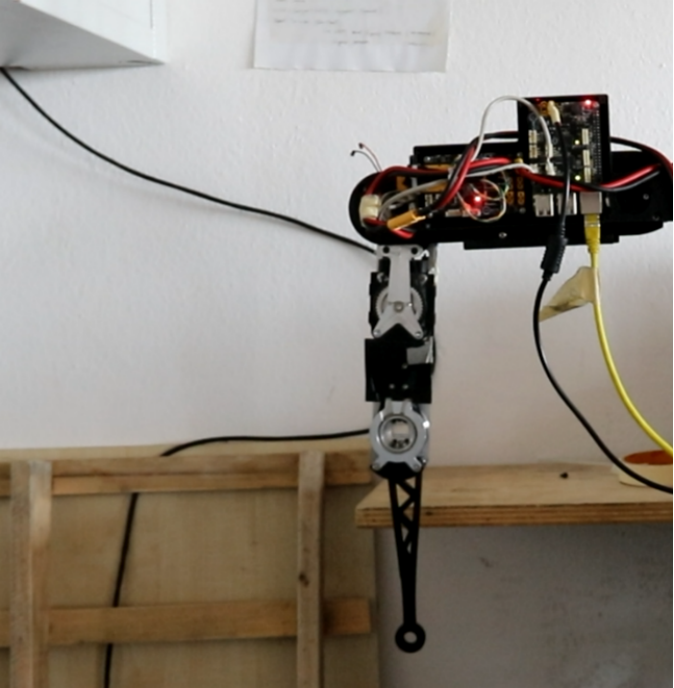}
    \includegraphics[width=0.24\linewidth]{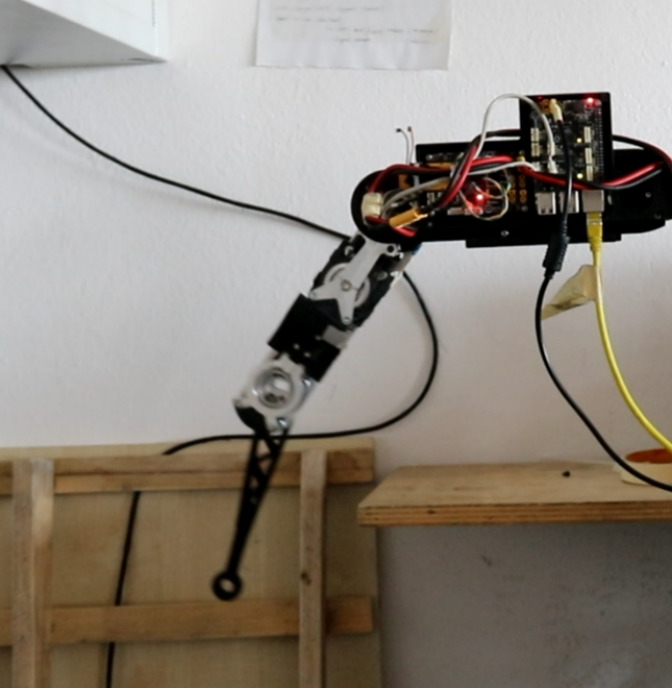}
    \includegraphics[width=0.24\linewidth]{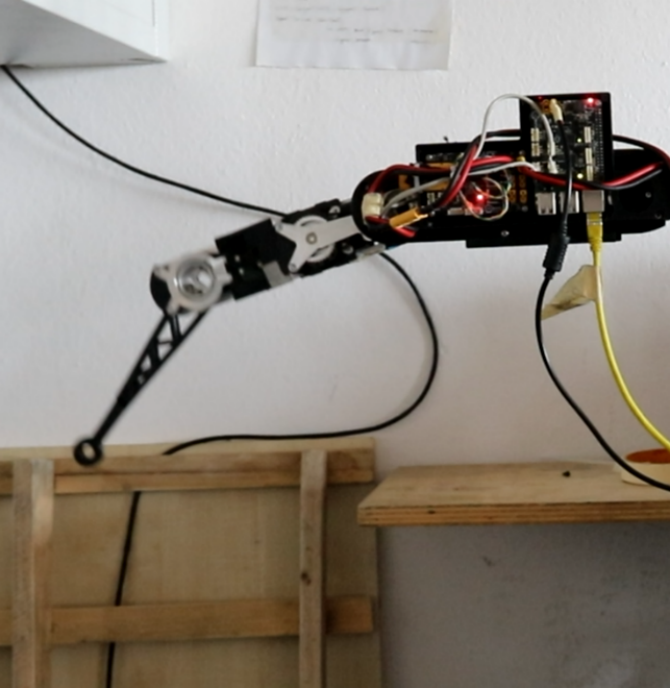}
    \includegraphics[width=0.24\linewidth]{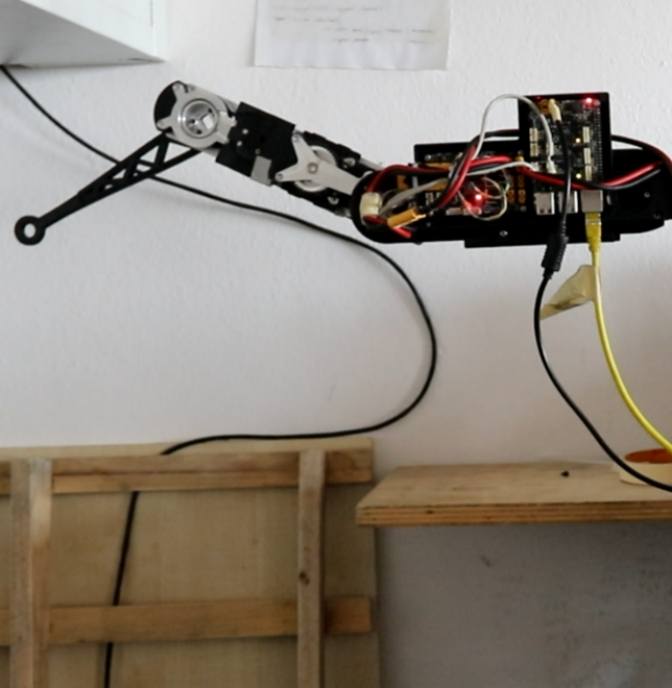}
    \caption{Photo-sequence of Reinforced Iterative Learning Control (RILC) on tracking trajectory with
    $y_{\textnormal{d}}(t_{\textnormal{f}})=[2.0,-1.0]^{\top}$.}
    \label{fig:frames}
\end{figure}

\paragraph{Test 1}
To compare the effectiveness of RILC \wrt ILC, the trajectory following task is conducted for the four target configurations, e.g., $y_{\textnormal{d}} (t_{\textnormal{f}})=\{[1.0,\,-1.0]^\top, [1.0,\,1.0]^\top,$ $[2.0,\,-1.0]^\top, [\pi/2,\,\pi/2]^\top\}$. The tracking performances are shown in Fig. \ref{fig:crossover}. 
We analyze the transient convergence phase using a first-order exponential model. For ILC, the RMSE evolution is modeled as $\bar\varepsilon_j \approx \bar\varepsilon_0 \bar\Omega^{j}$, and for RILC as $\varepsilon_j \approx \varepsilon_0 \Omega^{j}$, where $\bar\varepsilon_j$ and $\varepsilon_j$ denote the tracking RMSE at iteration~$j$. With a slight abuse of notation, we employ the same symbols $\Omega$ and $\bar\Omega$ as in Proposition~\ref{prop:RILCvsILC} to denote their empirical estimates. We identify the initialization gain $G_0 = \bar\varepsilon_0 / \varepsilon_0$ and the contraction ratio $\kappa = \Omega / \bar\Omega$, quantifying the zero-shot error reduction and the relative convergence rate. If $\kappa < 1$, the two exponential curves intersect at a finite crossover iteration $j^* \approx \log G_0 / \log \kappa$, defining the boundary within which RILC's initialization advantage dominates. The model is fitted via log-linear regression over an adaptive transient window, as discussed in Section~\ref{sec:disc}. The results are shown in Fig.~\ref{fig:crossover} and Table~\ref{tab:crossover}. 
The control contributions for the case $y_{\textnormal{d}}(t_{\textnormal{f}})=[2.0,-1.0]^{\top}$ are illustrated in Fig.~\ref{fig:contrib_torque}, and the tracking output is shown in Fig.~\ref{fig:track_}. Finally, Fig.~\ref{fig:frames} shows a photo-sequence of reaching using RILC.

\begin{figure}
    \centering
        \subfloat[\label{fig:frame_disturb}]{\includegraphics[width=0.22\linewidth]{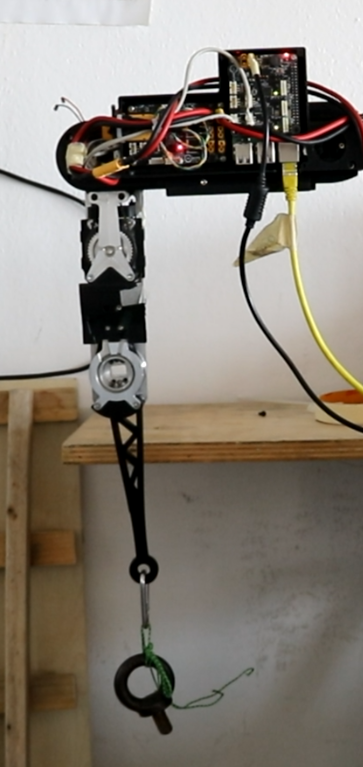}}
    \quad
        \subfloat[\label{fig:rmse_disturb}]{\includegraphics[width=0.6\linewidth]{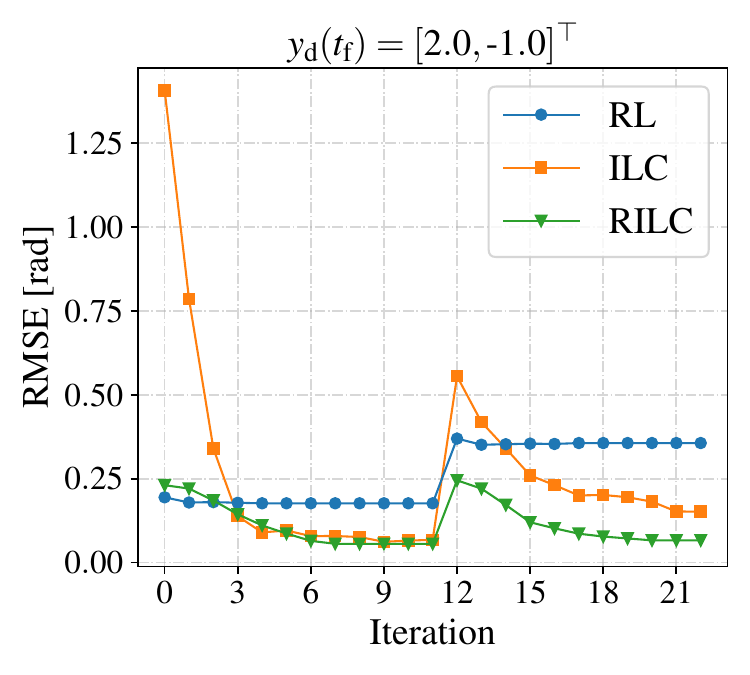}}
    \caption{Fig.~\ref{fig:frame_disturb} is the photo configuration of the disturbance applying. Fig.~\ref{fig:rmse_disturb} shows RMSE over iteration, a disturbance is applied to the robot in iteration $11$-th. Comparison of Reinforcement Learning (RL), Iterative Learning Control (ILC), and Reinforced Iterative Learning Control (RILC), approaches for the desired trajectory with $y_{\textnormal{d}}(t_{\textnormal{f}}) = [2.0,-1.0]^{\top}$. }
\end{figure}

\begin{figure}
    \centering
    \includegraphics[width=0.85\linewidth, trim={2cm 0 2cm 0}, clip]{
        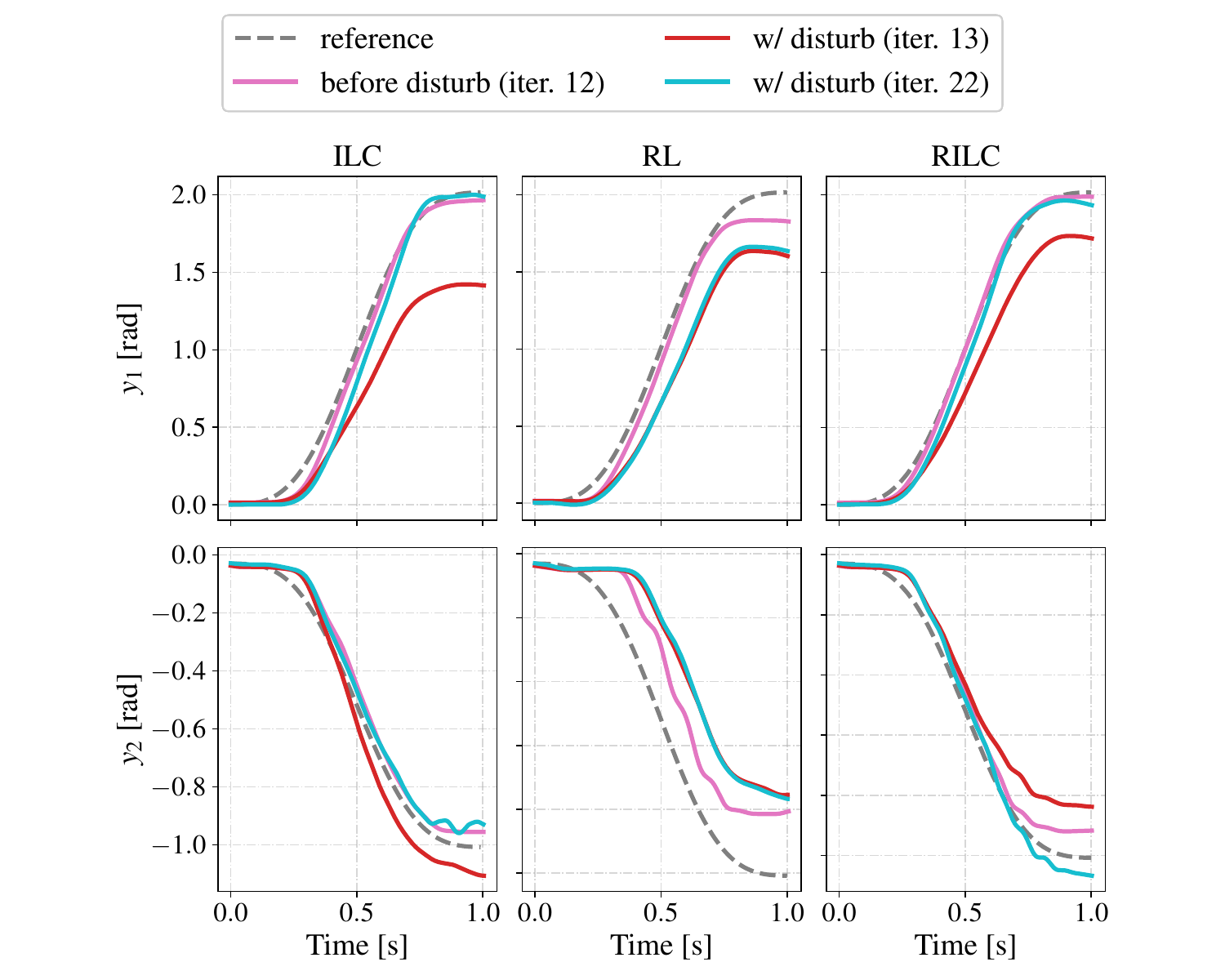
    }
    \caption{Output tracking evolution for desired trajectory with
    $y_{\textnormal{d}}(t_{\textnormal{f}}) = [2.0,-1.0]^{\top}$, without and with (w/)
    disturbance, in iterations $12$-th, $13$-th, $22$-nd. Comparison of Reinforcement Learning (RL), Iterative Learning Control (ILC), and Reinforced Iterative Learning Control (RILC), approaches.}
    \label{fig:disturb_track_}
\end{figure}

\paragraph{Test 2}
We tested the tracking performance in the presence of a unmodeled change in the dynamics of the robot , to validate the adaptability of proposed framework in the presence of such disturbances, Fig. \ref{fig:frame_disturb}, during trajectory following with $y_{\textnormal{d}} (t_{\textnormal{f}})=[2.0,\, -1.0]^{\top}$. The disturbance is simulated by an additional weight of $m=0.2\, \si{\kilogram}$ applied at the tip of the robot. Note that the additional mass is not modeled in the dynamics of the system and is around 20\% of the total mass of the robot. These tests are conducted with different controllers, \ie, RL, ILC, and RILC, to highlight the robustness of the control employed. The tracking performance is shown in Fig. \ref{fig:rmse_disturb}. The tracking outputs evolution before the application of disturbance, when it occurs and the new convergence, is illustrated in Fig.~\ref{fig:disturb_track_}.

\begin{table}[ht]
\caption{Crossover analysis for the real-robot experiments.
         Parameters $\bar\varepsilon_0$ and $\varepsilon_0$ are the initial tracking errors for ILC and RILC, respectively; $\bar\Omega$ and $\Omega$ are the corresponding fitted contraction rates; and $j^*$ is the crossover iteration.}
\label{tab:crossover}
\centering
\small
\begin{tabular}{lccccc}
\toprule
Target & $\bar\varepsilon_0$ & $\varepsilon_0$
        & $\bar\Omega$ & $\Omega$
        & $j^*$ \\
\midrule
$[1.0,\,-1.0]^\top$ & 0.660 & 0.140 & 0.666 & 0.812  & 7.9 \\
$[1.0,\,1.0]^\top$   & 0.682 & 0.063 & 0.762 & 0.973 & 9.8 \\
$[2.0,\,-1.0]^\top$  & 0.885 & 0.261 & 0.625 & 0.794 & 5.1 \\
$[\pi/2,\,\pi/2]^\top$ & 0.989 & 0.094 & 0.730 & 0.884 & 12.2 \\
\bottomrule
\end{tabular}
\end{table}


    \section{Discussion}
\label{sec:disc}

In this section, we discuss the results presented in Sec. \ref{sec:val}.

The RILC framework shows good tracking performance in any test, achieving the Goals of this work. Regarding the simulation results, Fig. \ref{fig:tb_err}, shows the benefits of the concurrent application of ILC during the training. 
Indeed, the training of RILC, after the initial phase, reaches a higher reward value than the RL baseline. Further, the comparison in Fig. \ref{fig:tb_err} indicates that RILC yields a more precise tracking trajectory under the same reward function.
It is worth noting that although the reward function increased faster in the initial phase of baseline RL it is not sufficient to obtain a precise tracking performance.
Conversely, the use of ILC helps to maximize the reward functions to achieve good tracking. 
Tab.~\ref{tab:resum_train} and Fig.~\ref{fig:color_map} evidence a stronger generalization capability \wrt RL, also when facing trajectories not used in the training, demonstrating that the high-quality samples generated by the RILC loop contribute to learning a more robust policy. Moreover, the proposed framework is able to improve performance over subsequent trials, in contrast to the static deployment of the baseline RL policy.

Table~\ref{tab:resum_train} enables a comprehensive comparison of all methods, namely RILC, RL, ILC, NOILC, and C-ILC. At iteration~0, RILC provides the best zero-shot initialization with a median error of 0.182 rad, significantly lower than the other baselines. At convergence in nominal conditions, NOILC achieves the lowest error floor, followed by ILC, while RILC plateaus at 0.018 due to the approximation nature of the neural policy. C-ILC, thanks to its pre-trained basis, achieves a lower median error at iteration~0 than ILC and NOILC, though higher than RILC. Its performance remains similar across nominal and mismatched conditions, reflecting the robustness of the basis functions to unmodeled dynamics, while its asymptotic convergence in the nominal case remains lower than ILC and NOILC. Under mismatched conditions, instead, RILC converges to the best asymptotic optimum of 0.063, outperforming all other methods, followed by C-ILC. These results highlight that RILC trades marginal nominal asymptotic accuracy for superior rapid adaptation and robustness against model mismatches.

Finally, the real-world experiments presented in Figs.~\ref{fig:crossover} and \ref{fig:rmse_disturb} confirm the effectiveness of the approach.
The proposed RILC framework shows general superior performance in terms of RMSE compared to ILC for all trajectories tested while maintaining convergence over iterations, see Fig. \ref{fig:crossover}, showing that G1) and G2) are achieved.
Fig.~\ref{fig:crossover} and Table~\ref{tab:crossover} report the crossover analysis, where the RMSE evolution is characterized via a first-order exponential model as $\varepsilon_j \approx \varepsilon_0 \Omega^{j}$. While this model assumes convergence to zero error, it provides a useful local approximation of the learning dynamics during the transient phase before saturation, confirmed by coefficients of determination $R^2 > 0.98$ for ILC and $R^2 > 0.93$ for RILC. The analysis identifies an initialization gain $G_0 \in [3.4, 10.9]$, confirming superior tracking from the very first trial. The first-order models predict a crossover at $j^* \in [5.1, 12.2]$, where ILC would overtake RILC under ideal convergence. However, this crossover consistently falls beyond the saturation horizon, where the exponential approximation breaks down. Since both controllers reach their practical noise floor before $j^*$, the crossover is never physically realized and RILC maintains a lower tracking error across all iterations. The higher contraction rates $\Omega$ reported in Table~\ref{tab:crossover} do not indicate architectural limitations but follow from the initialization gain reducing the available margin for iterative improvement, since RILC starts closer to the asymptotic floor.
This physical interpretation is further supported by the tracking profiles in Fig.~\ref{fig:track_}, where the first attempt for joint 1 is quite similar to the final one, while joint 2 has lower performance. However, both show improved tracking performance compared to ILC.
This performance mismatch is probably due to the gap between the simulated and real models, where joint 2 suffers from complex mechanical coupling and nonlinear friction that are difficult to model perfectly in simulation.
Although this little degradation of tracking performance, the RILC compensates for it over iterations thanks to the collaborative intervention of ILC.

The effectiveness of the RL-ILC collaboration within the RILC framework is evident by analyzing the control contribution of Fig. \ref{fig:contrib_torque}. The RL policy provides a highly accurate feedforward initialization. Already at the first trial, the RL torque profile exhibits a correlation of 91.9\% and 87.0\% for joint 1 and 2, respectively, with the final total control signal reached at convergence. This high similarity allows RILC to maintain the feedback term stable and bounded below 0.15 Nm root mean square (RMS) throughout the entire learning process. In contrast, standalone ILC requires several iterations to reach such levels, starting with feedback efforts up to 0.467 Nm, i.e., over three times the RMS effort of RILC, to counteract unmodeled dynamics and uncertainties. 

Finally, Fig.~\ref{fig:rmse_disturb} and Fig.~\ref{fig:disturb_track_} demonstrate that the RL component provides superior robustness against unmodeled disturbances within the same iteration. In Test 2, the application of an unmodeled tip mass causes the peak RMSE of standalone ILC to jump to 0.398 rad. The RILC framework limits this peak to 0.245 rad, reducing the instantaneous impact by 38.4\%. Standalone RL maintains a steady tracking error without trial-to-trial improvement. While ILC structurally requires a subsequent trial to incorporate the error into its feedforward memory, RILC maintains a tighter performance floor throughout the disturbance event, confirming its effectiveness in tackling non-repetitive unmodeled dynamics.

    \section{Conclusion and future work}
\label{sec:concl}
In this work, we have proposed the Reinforced Iterative Learning Control (RILC) framework, which combines Iterative Learning Control (ILC) and Reinforcement Learning (RL), to address the precise trajectory tracking problem for compliant robots.
A key contribution of this work is the derivation of formal convergence guarantees for this combined architecture. This distinguishes RILC from existing approaches based on hyperparameter tuning or additive Residual RL, which often lack theoretical proofs or are limited to single-task robustness without addressing generalization. Unlike methods that decouple learning from execution, we proved that the online RL-ILC interaction is synergistic.
These theoretical guarantees provide the formal basis for addressing the generalization problem across multiple desired trajectories.

The numerical and experimental validation on a 4-DoFs underactuated manipulator confirmed the effectiveness of this approach compared to standalone RL, ILC, NOILC, and C-ILC baselines. The results demonstrate that RILC achieves high tracking performance from the very first trial of a new trajectory, effectively overcoming the initialization limitations of standard ILC and NOILC, while maintaining the ability to refine precision over subsequent iterations. Furthermore, the coupling enables ILC to provide high-quality samples that guide the RL training, while the RL policy generalizes across the workspace, outperforming multi-reference ILC approaches on out-of-distribution trajectories.
Finally, the framework exhibited superior robustness against unmodeled disturbances and model mismatches, confirming that the collaborative intervention of ILC and RL can effectively bridge the reality gap that each method alone struggles to overcome. These results validate the theoretical guarantees derived in this work, showing that the concurrent RL-ILC coupling preserves convergence while providing cross-trajectory generalization and enhanced robustness.

This work provides a foundation for future optimizations, such as studying the impact of different RL cost functions on the ILC convergence rate. Having demonstrated the
convergence and effectiveness of this benchmark, the natural next step is the application of RILC to more complex dynamic interaction environments.

    \appendix 
    \section{Link-side tracking simulation}
\label{sec:sim_link}

\begin{figure}[htbp]
    \centering
    \subfloat[\label{fig:track_sim_q}]{\includegraphics[width=0.47\linewidth]{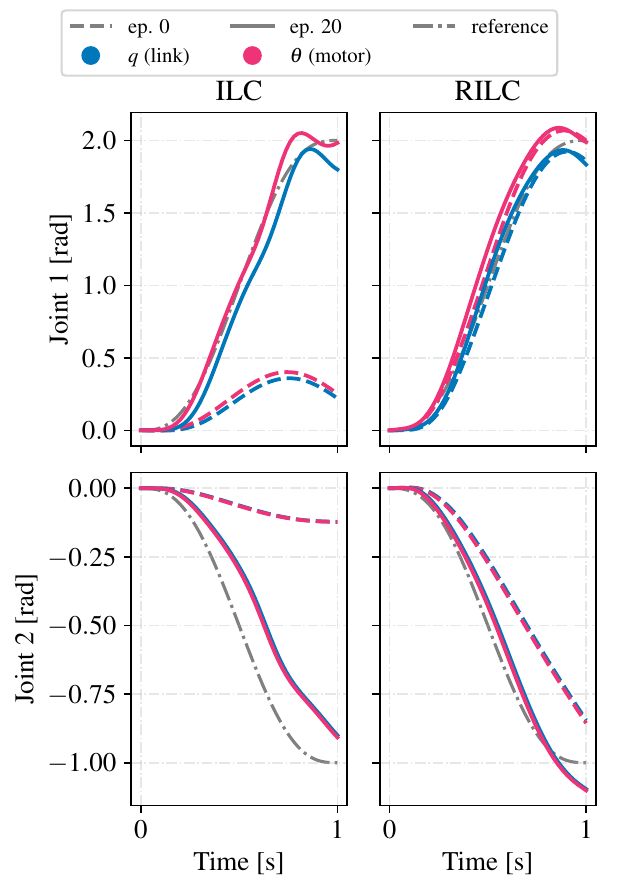}} \quad
    \subfloat[\label{fig:rmse_sim_q}]{\includegraphics[width=0.4\linewidth]{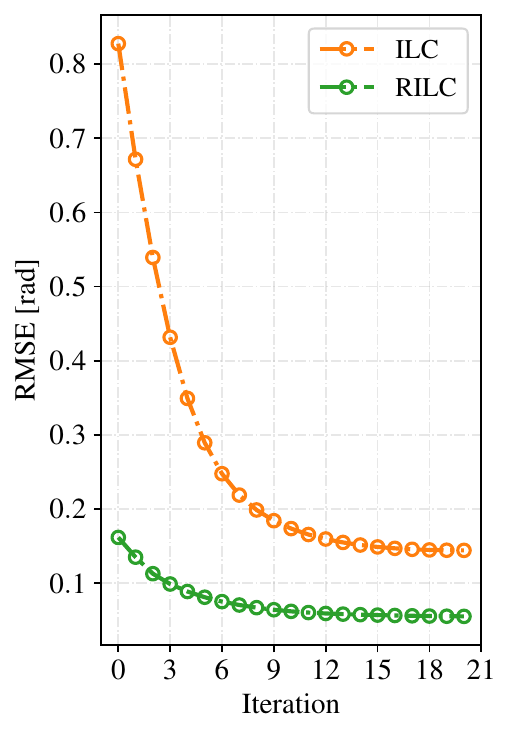}}
    \caption{Simulation results comparison between ILC and RILC for link-side position tracking $y(t)=q(t)$ for final position $y_{\textnormal{d}}(t_{\textnormal{f}}) = [2.0, -1.0]^{\top}$. (a) Motor and link position trajectories. (b) RMSE evolution over iterations.}
    \label{fig:link_side_sim}
\end{figure}

To evaluate the theoretical generality of the RILC architecture, we conduct a simulation study for link-side position tracking $y(t)=q(t)$ with a target final position $y_{\textnormal{d}}(t_{\textnormal{f}}) = [2.0, -1.0]^{\top}$. This non-collocated configuration yields a higher relative degree and introduces non-minimum phase zero dynamics, increasing control complexity. Despite this, RILC achieves stable convergence and provides a significantly tighter convergence than standalone ILC, as illustrated in Fig.~\ref{fig:link_side_sim}. Fig.~\ref{fig:track_sim_q} shows the distinct evolution of the motor and link positions, due to the joint's elastic dynamics. The ILC update law uses diagonal learning matrices $\chi_0 = 0.165 I_2$, $\chi_1 = 0.2 I_2$, $\chi_2 = 0.00115 I_2$, with higher-order gains set to zero. Fig.~\ref{fig:track_sim_q} reports the tracked link position, while Fig.~\ref{fig:rmse_sim_q} shows the RMSE progression over trials.

\section{Baseline implementation details}
\label{app:baselines}

This section details the implementation of the baseline controllers used in the simulation benchmarks of Section~\ref{sec:val}.

\subsection{Norm-Optimal ILC (NOILC)}
\label{app:noilc}

The NOILC baseline \cite{amann1996iterative} is implemented with a trial-domain cost function
\begin{equation}
    J_j = \sum_{t=0}^{t_{\textnormal{f}}} \bigl[ e_j(t)^{\top} Q e_j(t) + \delta\bar{u}^\textnormal{N}_j(t)^{\top} R \,\delta\bar{u}^\textnormal{N}_j(t) \bigr],
\end{equation}
where $\delta\bar{u}^\textnormal{N}_j(t) = \bar{u}^\textnormal{N}_j(t) - \bar{u}^\textnormal{N}_{j-1}(t)$. The weighting matrices are set to $Q = R = I$. The lifted-domain update law is
\begin{equation}
    \bar{u}^\textnormal{N}_{j+1} = \bar{u}^\textnormal{N}_j + \bigl[(G^{\top} Q G + R)^{-1} G^{\top} Q\bigr] e_j,
\end{equation}
where $G$ denotes the lifted system matrix obtained by linearizing the robot dynamics along the desired trajectory.

\subsection{Combined ILC (C-ILC)}
\label{app:cilc}

We adopt the C-ILC framework of \cite{tsurumoto2023task}. The C-ILC law is
\begin{equation}
    u_j^\textnormal{C} = u_j^w + \bar{u}_j \,,
\end{equation}
where $\bar{u}_j$ is defined in \eqref{eq:eq_ilc}, $u_j^w = \Phi_0\, w_j$ is the basis term and $\Phi_0$ is built from the sampled trajectory evolution using
\begin{equation}
    \varphi(y) = [ \ddot{y}_1,\; \ddot{y}_2,\; \dot{y}_1,\; \dot{y}_2,\;
                      \sin(y_1),\; \sin(y_2),\; 1 ]^\top .
\end{equation}
The iterative update adapts $w$ and $\bar{u}_j$ per trial
\begin{align}
    w_{j+1} &= w_j + \Phi_0^\dagger (e_j + \bar{u}_j ) \,, \label{eq:cilc_w} \\
    \bar{u}_{j+1} &= \bar{u}_j + \chi_i e_j^{(i)} +  (u_j^w - u_{j+1}^w) \,, \label{eq:cilc_ilc}
\end{align}
where $\Phi_0^\dagger = (\Phi_0^\top\Phi_0 + \lambda I)^{-1}\Phi_0^\top$ is the regularized pseudoinverse. The correction term $(u_j^w - u_{j+1}^w)$ ensures the total feedforward evolves coherently when $w$ updates, \cite{tsurumoto2023task}. We adopt $\chi_0 = 0.4$, $\chi_1 = 0.2$, and $\lambda = 0.1$.

To enable zero-shot transfer, we pre-train $w$ via batch gradient descent on the minimum-jerk trajectories of the reduced workspace, see Section~\ref{sec:sim_res}, stacking the basis matrix $\mathcal{F}$, tracking error $\mathcal{E}$, and ILC residual $\bar{U}$ across all training trajectories. The batch update is
\begin{equation}
    w \leftarrow w + \gamma_w \, \mathcal{F}^\dagger (\mathcal{E} + \bar{U}) \,,
\end{equation}
with learning rate $\gamma_w = 0.6$, repeated until convergence.

    \printcredits

    \section*{Declaration of competing interest}
    The authors declare that they have no known competing financial interests or personal relationships that could have appeared to influence the work reported in this paper.

    \bibliographystyle{cas-model2-names}
    \bibliography{cas_refs}

\end{document}